\documentclass[11pt]{article}

\usepackage[final]{acl}

\usepackage{times}
\usepackage{latexsym}
\usepackage{booktabs,multirow,colortbl,xcolor,longtable}
\usepackage[T1]{fontenc}

\usepackage[utf8]{inputenc}
\usepackage{amsmath}
\usepackage{float}
\usepackage{microtype}
\usepackage{array}
\usepackage{pgfplots}
\pgfplotsset{compat=1.18}
\usepgfplotslibrary{fillbetween}

\usepackage{inconsolata}

\usepackage{graphicx}
\usepackage{pdfpages}

\usepackage{afterpage}
\usepackage{makecell}

\title{Assessing the Impact of Language Disparity on Multilingual Linguistic Ability in Large Language Models}

\author{Zhanyu Chen \\
  University of Groningen \\
  \texttt{z.chen.67@student.rug.nl} \\\And
  Jaap Jumelet \\
  University of Groningen \\
  \texttt{j.w.d.jumelet@rug.nl} \\}

\begin{document}
\maketitle
\begin{abstract}
Claims about the grammatical competence of multilingual language models vary sharply with how competence is measured, yet the interaction between evaluation paradigm, post-training, and language resource availability has not been systematically examined. 
We evaluate base and post-trained models from six families on MultiBLiMP, a syntactic minimal-pair benchmark covering 101 languages, using four evaluation methods.
We report three principal findings. 
First, post-training degrades grammatical competence, but the magnitude of this effect is reduced unevenly by model scale, while low-resource languages bear the highest cost. 
Second, post-trained models retain grammatical knowledge they cannot articulate through explicit prompting, yet this  is measurable only in high-resource languages, because near-chance baselines in low-resource settings leave little knowledge to hide. 
Third, native-language prompting recovers otherwise hidden competence on low-resource languages, demonstrating that only high-resource languages can be probed directly from unprompted probabilities. 
We conclude that multilingual grammatical evaluation must adopt language-informed, multi-paradigm protocols to avoid systematically underestimating low-resource abilities.
\end{abstract}

\section{Introduction}
Multilingual Large Language Models (LLMs) have demonstrated remarkable capabilities in processing and generating text across a wide range of languages.
Scaling LLMs up multilingually has led to an increase of evaluation benchmarks that evaluate model performance across a broader spectrum of languages, taking fine-grained linguistic and cultural properties into account \citep{chang2025globalpiqaevaluatingphysical, Vayani_2025_CVPR}.
These benchmarks have exposed that language disparity persists in NLP, with advanced LLM abilities remaining limited to a small subset of high-resource languages \citep{blasi-etal-2022-systematic,wendler-etal-2024-llamas}.

While various works have investigated the impact of language disparity on downstream tasks \citep{qi-etal-2023-cross, xuan-etal-2025-mmlu}, a detailed investigation of its impact on \textit{linguistic ability} remains lacking.
For this, we leverage {MultiBLiMP} \citep{jumelet-etal-2026-multiblimp}, a comprehensive multilingual benchmark consisting of linguistic minimal pairs across 101 languages. 
However, the exact way of measuring linguistic ability in LLMs remains a topic of discussion \citep{10.1162/TACL.a.611}, and we therefore measure model performance across a range of metrics.

We conduct detailed regression analyses to understand performance differences at the language level, focusing on two previously underexplored sources of disparity.
First, we investigate the \textit{post-training} phase, in which pre-trained LLMs are fine-tuned for instruction following and human preference alignment.
Due to the scarcity of language-specific post-training data, this stage predominantly uses high-resource languages \citep{Longpre2024}, potentially degrading low-resource language abilities \citep{kew-etal-2024-turning}.
We quantify this degradation through a \textbf{linguistic alignment tax}, measuring the performance gap between base and post-trained models.
Our analysis reveals that this tax is not determined by model scale, but rather by factors related to post-training recipe, although the specifics of that require further investigation. Crucially, low-resource languages pay substantially higher costs than high-resource ones.
 
Second, building on \citet{song2025languagemodelsfailintrospect}'s finding that LLMs fail to ``introspect'' about their grammatical knowledge in English, we extend this analysis across all 101 languages in MultiBLiMP.
We introduce the \textbf{inarticulate gap}, measuring the difference between direct grammaticality judgments and meta-linguistic evaluation.
Our results show that post-trained models retain grammatical knowledge they cannot articulate through explicit prompting, but critically, this gap is only measurable in high-resource languages, where performance exceeds chance levels.
In low-resource settings, near-chance baselines obscure any hidden competence.
Statistical analysis shows the alignment tax and inarticulate gap to be moderately negatively correlated in four of the six families.
 
Finally, we demonstrate that native-language prompting, in which we provide translated instructions in the target language itself, yields disproportionate gains for low-resource languages, revealing grammatical knowledge that remains weakly expressed under direct probability assessment.
These findings stress the need for language-informed, multi-paradigm evaluation protocols that avoid systematically underestimating low-resource language capabilities.\footnote{Code, evaluation outputs, and analysis notebooks are available at \url{https://github.com/EricZhanyuChen/multiblimp_eval}.}

\section{Related Work}
\subsection{Evaluating Linguistic Ability}
The grammatical competence of language models has commonly been 
assessed through the minimal pair paradigm, where a model is 
considered correct if it assigns higher log-probability to a 
grammatical sentence over a minimally perturbed ungrammatical 
counterpart \citep{linzen-etal-2016-assessing, marvin-linzen-2018-targeted, 
warstadt-etal-2020-blimp-benchmark}. A parallel line of work evaluates 
grammaticality through metalinguistic prompting, asking models directly 
whether a sentence is grammatical. \citet{doi:10.1073/pnas.2309583120} 
found near-chance accuracy and a systematic yes-response bias under 
this paradigm, but \citet{hu-levy-2023-prompting} argue that such 
failures reflect auxiliary task demands rather than grammatical 
incompetence, since answering metalinguistic questions requires 
instruction-following ability independent of grammatical knowledge. 
\citet{doi:10.1073/pnas.2400917121} provide supporting evidence, 
showing that log-probability comparisons reveal substantially stronger 
alignment between models and human judgments.

\citet{song2025languagemodelsfailintrospect} directly compare both 
paradigms on the same minimal pairs, finding that post-trained 
models retain grammatical knowledge they fail to articulate through 
the chat interface---the \textit{inarticulate gap}. However, this 
line of work is conducted exclusively in English, leaving open whether 
the same dissociation holds cross-lingually and how it interacts with 
language resource availability and post-training.

\subsection{Multilingual Evaluation}

Minimal pair benchmarks have since been extended to languages beyond 
English, including Chinese \citep{liu2025systematicassessmentlanguagemodels}, Russian 
\citep{taktasheva-etal-2024-rublimp}, Dutch \citep{suijkerbuijk-etal-2025-blimp}, 
Japanese \citep{someya-oseki-2023-jblimp}, and Icelandic 
\citep{armannsson-etal-2025-icelandic}. Cross-lingual evaluation has 
been attempted at small scale by \citet{gulordava-etal-2018-colorless} 
and \citet{mueller-etal-2020-cross}, and is scaled to 101 languages by 
\citet{jumelet-etal-2026-multiblimp} with MultiBLiMP. 
However, all of these efforts evaluate exclusively with log-probability, 
leaving the effect of evaluation method unexamined across languages.

The choice of prompt language adds a further complication. 
\citet{etxaniz-etal-2024-multilingual} show that multilingual models 
perform better when prompted in English than in the target language, 
and \citet{poelman-lhoneux-2025-roles} argue that using English prompts 
to evaluate non-English grammatical knowledge conflates task performance 
with language understanding. No existing work simultaneously applies 
multiple evaluation methods across a typologically diverse language set. 
We address this gap by comparing four evaluation methods across 101 languages spanning four resource tiers.

\section{Methods}
\subsection{Evaluation Methods}
We utilize the Language Model Evaluation Harness (\texttt{lm-eval}) framework \citep{eval-harness} to standardize our assessment. 
For minimal pair evaluation, which consists of a grammatical sentence ($S_{cor}$) and an ungrammatical sentence ($S_{inc}$), we evaluate the model's ability to assign a higher probability to the correct sentence. 
To investigate the impact of post-training and the discrepancy between intrinsic knowledge and instruction following and model's preference on English, we implemented four distinct evaluation strategies, categorized into Discriminative and Meta paradigms.

\paragraph{Discriminative Evaluation (Likelihood-based)}
In these methods, the model acts as a probability estimator for the linguistic sequences themselves. We compute the log-likelihood of both the correct ($S_{cor}$) and incorrect ($S_{inc}$) sentences conditioned on a specific prefix context $C$. The model is considered correct if $P(S_{cor} | C) > P(S_{inc} | C)$. We define three variations of the prefix to test context sensitivity:

\begin{itemize}
    \item \textbf{Direct Probability (Zero-Shot):} This method assesses the model's knowledge without any prompt engineering: no instruction is prepended, and the model scores the sequence probability of the two sentences on their own. 
    Base models are scored on raw text and post-trained models on the same sentences within their own chat template, so this condition reflects each model type's standard inference format rather than a single shared one.
    
    \item \textbf{English-Prompted Probability:} To evaluate cross-lingual alignment through English-centric priming, we prepend a standardized English context string $C_{eng}$ (e.g., \textit{``The following is a sentence in [Language].''}) before the target sentences. This tests whether English instructions can successfully trigger the retrieval of low-resource language knowledge.
    
    \item \textbf{Native-Prompted Probability:} Similar to the English prompt, but the context string $C_{native}$ is provided in the target language itself. This evaluates the model's ability to process instructions and linguistic cues within the same low-resource language space, minimizing cross-lingual interference. For modern languages, $C_{native}$ is
a direct translation of the English template (e.g., \textit{``Dies ist ein Satz
auf Deutsch.''} for German), generated by Gemini 3.0 Pro and verified by
Claude 4.6 Opus. For 14 languages the full template cannot be rendered: for twelve of them (e.g., Gothic,
Hittite, Church Slavonic) the cue is shortened to a statement naming the language itself,
and for Egyptian and Phrygian an attested phrase in that language is used instead. The
complete list is provided in Appendix~\ref{sec:appendix_native_prompts}.
\end{itemize}

\paragraph{Meta Evaluation}
This method shifts the task from sequence modeling to explicit instruction following. 
We restructure the minimal pair task into a Question-Answering (QA) format. The model is presented with a chat-style prompt: \textit{``Which of the following sentences is grammatically correct in [Language]? Only respond with 1 or 2.''}. 
The position of the grammatical sentence is not fixed: it is assigned to option 1 or option 2 according to the parity of its character length, which balances the two positions across the benchmark and rules out a positional option bias.
Unlike the discriminative methods that score the full sentence, here the model must explicitly generate a classification token. 
This method specifically tests the model's ability to adhere to formatting constraints and explicitly articulate its linguistic knowledge, and has been referred to as `\textit{meta-linguistic} evaluation' in prior work \citep{song2025languagemodelsfailintrospect}.

\paragraph{Metric}
For all methods, we report Accuracy as the primary metric, measuring the percentage of instances where the model correctly assigns higher probability to the grammatical sentence (in discriminative evaluation) or generates the correct option label (in meta evaluation). Accuracy is averaged equally across languages, so that each language contributes the same weight regardless of the number of test items it contains.
Because languages contribute equally regardless of test-set size, we quantify the resulting sampling noise directly: propagating per-language binomial standard errors through the equal-weighted mean gives an aggregate standard error of at most 0.50 percentage points across all 48 cells of Table~\ref{tab:model_accuracy}. The largest values occur in the meta cells, whose accuracy is closest to chance.

\subsection{Data}
\paragraph{Evaluation Benchmark}
We evaluate our models on {\textbf{MultiBLiMP}} \citep{jumelet-etal-2026-multiblimp}.
This dataset facilitates targeted syntactic evaluation across {101 languages} covering a diverse set of language families including Indo-European, Uralic, Turkic, and Afro-Asiatic.

MultiBLiMP consists of 121,305 minimal pairs. Each pair comprises a grammatical sentence ($S_{cor}$) and a minimally perturbed ungrammatical counterpart ($S_{inc}$), created automatically by leveraging annotations from Universal Dependencies and UniMorph. 
The dataset focuses specifically on morpho-syntactic agreement phenomena. Each minimal pair is meaning-matched: the ungrammatical variant differs from the grammatical one by a single morphological inflection. 
Specifically, it covers:
\begin{itemize}
    \item \textbf{Subject-Verb Agreement}: Tests agreement in number, person, and gender between the subject and the finite verb.
    \item \textbf{Subject-Participle Agreement}: Tests agreement features on participles, which is prominent in Romance and Slavic languages.
\end{itemize}

\paragraph{Resource Stratification: Common Crawl}
To investigate the correlation between model alignment and pre-training data volume, we utilize language frequency statistics derived from the common crawl corpus, provided by \citet{NEURIPS2024_1e605762}.
These language statistics serve as a general proxy to language distributions in web-scale corpora: training data has not been released for any of the models we tested.
Based on the natural power-law distribution of web data, we categorize the languages into four distinct resource tiers:\vspace{0.2cm}

{\small
\begin{tabular}{@{}lrrl@{}}
\toprule
\textbf{Tier} & \textbf{CC Tokens} & \textbf{\#Lang} & \textbf{Examples} \\
\midrule
Super High & $>10^{11}$          & 7  & eng, rus, spa \\
High       & $10^{9}$--$10^{11}$ & 38 & por, nld, tur \\
Mid        & $10^{7}$--$10^{9}$  & 20 & guj, cym, bre \\
Low/Zero   & $\leq 10^{7}$       & 36 & abk, got, yrl \\
\bottomrule
\end{tabular}}\vspace{0.2cm}

These boundaries follow the natural power-law distribution of web data, where order-of-magnitude intervals correspond to qualitative shifts in resource availability. The resulting tier sizes reflect the skewed real-world distribution of language resources, with a small number of dominant languages and a long tail of low-resource ones.

\subsection{Models}
We evaluate models from six open-weight LLM families, each providing paired \textit{base} and post-trained (\textit{PT}) checkpoints for multiple model sizes. 
For models with native reasoning capabilities, the thinking mode is disabled during evaluation to ensure comparability across all model families.
Inputs to PT models are constructed using \texttt{apply\_chat\_template} for all evaluation methods.


\paragraph{Gemma~3}
Gemma~3 \citep{gemmateam2025gemma3technicalreport} is a dense decoder-only transformer trained on a mixture of over 140 languages. We evaluate four parameter scales (1B, 4B, 12B, 27B), each available in both a base and PT 
checkpoint. 

\paragraph{Llama~3.1/3.2}
Llama~3.1/3.2 \citep{grattafiori2024llama3herdmodels} is a family of dense decoder-only 
transformers. We evaluate three parameter scales (1B, 3B, 8B), where the 1B and 3B models are drawn from the Llama~3.2 release and the 8B model from Llama~3.1. Both releases officially support 8 languages; however, Meta notes that the models have been trained on a broader multilingual corpus than these 8 languages, making evaluation on the full MultiBLiMP benchmark feasible.

\paragraph{Qwen~3}
Qwen~3 \citep{yang2025qwen3technicalreport} is pre-trained on approximately 36 trillion tokens spanning 119 languages and dialects. The smaller variants (0.6B, 1.7B, 4B, 8B, 14B) use a dense decoder-only architecture; the 30B variant uses a sparse mixture-of-experts architecture with 3B active parameters per forward pass. We evaluate all six sizes. 

\paragraph{Qwen~3.5}
Qwen~3.5 \citep{qwen3.5} is a multimodal model family supporting 201 languages and dialects. The smaller variants (0.8B, 2B, 4B, 9B) use a dense transformer architecture; the 35B variant uses a sparse mixture-of-experts (MoE) architecture with 3B active parameters per forward pass. We evaluate all five sizes.

\paragraph{Gemma~4}
Gemma~4 \citep{gemma4_noyan2026} is the newest model in the Gemma family, including both dense (E2B, E4B, 31B) and mixture-of-experts (26B-A4B) architectures, with effective parameter counts of 2.3B, 4.5B, 30.7B, and 3.8B active (25.2B total), respectively. We evaluate all four sizes across both base and post-trained checkpoints.

\paragraph{TinyAya}
TinyAya \citep{salamanca2026tinyayabridgingscale} is a 3.35B-parameter dense decoder-only transformer developed by Cohere Labs, pre-trained on data spanning over 70 languages with a deliberate emphasis on low-resource language representation. Unlike the other model families evaluated here, TinyAya is specifically designed for multilingual coverage under constrained compute, making it a distinctive point of comparison for low-resource languages on MultiBLiMP. We only evaluate the base and the post-trained TinyAya-Global variant.

\begin{table}[t]
\centering
\scriptsize
\begin{tabular}{@{}ll>{\centering\arraybackslash}p{0.8cm}>{\centering\arraybackslash}p{0.8cm}>{\centering\arraybackslash}p{0.8cm}>{\centering\arraybackslash}p{0.8cm}}
& & \multicolumn{4}{c}{\textit{Evaluation Method}} \\ \arrayrulecolor{lightgray}\cmidrule(lr){3-6}\arrayrulecolor{black}
 & & \textbf{direct} & \textbf{meta} & \textbf{\makecell{native \\ prompt}} & \textbf{\makecell{English \\ prompt}} \\
\midrule
\multirow{2}{*}{Gemma~3 (27B)} & base & \cellcolor[rgb]{0.413,0.666,0.201}\makecell{84.17\\[-2pt]{\tiny$\pm$19.4}} & \cellcolor[rgb]{0.647,0.834,0.435}\makecell{73.17\\[-2pt]{\tiny$\pm$18.7}} & \cellcolor[rgb]{0.375,0.634,0.176}\makecell{86.21\\[-2pt]{\tiny$\pm$18.1}} & \cellcolor[rgb]{0.383,0.640,0.181}\makecell{85.77\\[-2pt]{\tiny$\pm$19.3}} \\
 & PT & \cellcolor[rgb]{0.507,0.743,0.266}\makecell{79.46\\[-2pt]{\tiny$\pm$18.9}} & \cellcolor[rgb]{0.630,0.823,0.414}\makecell{74.02\\[-2pt]{\tiny$\pm$17.9}} & \cellcolor[rgb]{0.467,0.711,0.235}\makecell{81.63\\[-2pt]{\tiny$\pm$17.5}} & \cellcolor[rgb]{0.498,0.737,0.255}\makecell{79.69\\[-2pt]{\tiny$\pm$17.1}} \\
\arrayrulecolor{lightgray}\cmidrule(lr){1-6}\arrayrulecolor{black}
\multirow{2}{*}{Gemma~4 (31B)} & base & \cellcolor[rgb]{0.383,0.640,0.181}\makecell{85.56\\[-2pt]{\tiny$\pm$19.1}} & \cellcolor[rgb]{0.406,0.660,0.196}\makecell{84.38\\[-2pt]{\tiny$\pm$16.1}} & \cellcolor[rgb]{0.329,0.595,0.147}\makecell{88.36\\[-2pt]{\tiny$\pm$16.9}} & \cellcolor[rgb]{0.337,0.602,0.152}\makecell{88.13\\[-2pt]{\tiny$\pm$17.9}} \\
 & PT & \cellcolor[rgb]{0.586,0.794,0.361}\makecell{76.09\\[-2pt]{\tiny$\pm$17.2}} & \cellcolor[rgb]{0.452,0.699,0.225}\makecell{82.33\\[-2pt]{\tiny$\pm$18.1}} & \cellcolor[rgb]{0.542,0.766,0.308}\makecell{78.06\\[-2pt]{\tiny$\pm$16.9}} & \cellcolor[rgb]{0.498,0.737,0.255}\makecell{79.77\\[-2pt]{\tiny$\pm$17.0}} \\
\arrayrulecolor{lightgray}\cmidrule(lr){1-6}\arrayrulecolor{black}
\multirow{2}{*}{Llama~3 (8B)} & base & \cellcolor[rgb]{0.475,0.718,0.240}\makecell{80.88\\[-2pt]{\tiny$\pm$18.4}} & \cellcolor[rgb]{0.939,0.965,0.900}\makecell{54.51\\[-2pt]{\tiny$\pm$9.9}} & \cellcolor[rgb]{0.429,0.679,0.211}\makecell{83.24\\[-2pt]{\tiny$\pm$19.2}} & \cellcolor[rgb]{0.444,0.692,0.220}\makecell{82.74\\[-2pt]{\tiny$\pm$18.7}} \\
 & PT & \cellcolor[rgb]{0.516,0.749,0.276}\makecell{79.19\\[-2pt]{\tiny$\pm$18.1}} & \cellcolor[rgb]{0.920,0.963,0.858}\makecell{57.31\\[-2pt]{\tiny$\pm$11.6}} & \cellcolor[rgb]{0.452,0.699,0.225}\makecell{82.07\\[-2pt]{\tiny$\pm$18.8}} & \cellcolor[rgb]{0.467,0.711,0.235}\makecell{81.39\\[-2pt]{\tiny$\pm$19.0}} \\
\arrayrulecolor{lightgray}\cmidrule(lr){1-6}\arrayrulecolor{black}
\multirow{2}{*}{Qwen~3 (30B)} & base & \cellcolor[rgb]{0.490,0.731,0.250}\makecell{80.37\\[-2pt]{\tiny$\pm$19.1}} & \cellcolor[rgb]{0.691,0.862,0.488}\makecell{71.40\\[-2pt]{\tiny$\pm$17.6}} & \cellcolor[rgb]{0.444,0.692,0.220}\makecell{82.56\\[-2pt]{\tiny$\pm$19.3}} & \cellcolor[rgb]{0.460,0.705,0.230}\makecell{81.73\\[-2pt]{\tiny$\pm$19.5}} \\
 & PT & \cellcolor[rgb]{0.533,0.760,0.297}\makecell{78.14\\[-2pt]{\tiny$\pm$18.6}} & \cellcolor[rgb]{0.746,0.893,0.565}\makecell{68.67\\[-2pt]{\tiny$\pm$17.1}} & \cellcolor[rgb]{0.490,0.731,0.250}\makecell{80.36\\[-2pt]{\tiny$\pm$18.7}} & \cellcolor[rgb]{0.498,0.737,0.255}\makecell{79.80\\[-2pt]{\tiny$\pm$19.4}} \\
\arrayrulecolor{lightgray}\cmidrule(lr){1-6}\arrayrulecolor{black}
\multirow{2}{*}{Qwen~3.5 (35B)} & base & \cellcolor[rgb]{0.429,0.679,0.211}\makecell{83.45\\[-2pt]{\tiny$\pm$19.0}} & \cellcolor[rgb]{0.460,0.705,0.230}\makecell{81.91\\[-2pt]{\tiny$\pm$17.9}} & \cellcolor[rgb]{0.390,0.647,0.186}\makecell{85.47\\[-2pt]{\tiny$\pm$18.8}} & \cellcolor[rgb]{0.383,0.640,0.181}\makecell{85.56\\[-2pt]{\tiny$\pm$19.0}} \\
 & PT & \cellcolor[rgb]{0.444,0.692,0.220}\makecell{82.44\\[-2pt]{\tiny$\pm$18.6}} & \cellcolor[rgb]{0.483,0.724,0.245}\makecell{80.48\\[-2pt]{\tiny$\pm$18.2}} & \cellcolor[rgb]{0.398,0.653,0.191}\makecell{84.89\\[-2pt]{\tiny$\pm$17.8}} & \cellcolor[rgb]{0.406,0.660,0.196}\makecell{84.59\\[-2pt]{\tiny$\pm$19.1}} \\
\arrayrulecolor{lightgray}\cmidrule(lr){1-6}\arrayrulecolor{black}
\multirow{2}{*}{TinyAya (3.35B)} & base & \cellcolor[rgb]{0.568,0.783,0.340}\makecell{76.68\\[-2pt]{\tiny$\pm$19.3}} & \cellcolor[rgb]{0.936,0.965,0.894}\makecell{55.01\\[-2pt]{\tiny$\pm$10.0}} & \cellcolor[rgb]{0.542,0.766,0.308}\makecell{78.06\\[-2pt]{\tiny$\pm$20.4}} & \cellcolor[rgb]{0.551,0.771,0.319}\makecell{77.51\\[-2pt]{\tiny$\pm$21.2}} \\
 & PT & \cellcolor[rgb]{0.621,0.817,0.403}\makecell{74.61\\[-2pt]{\tiny$\pm$18.2}} & \cellcolor[rgb]{0.936,0.965,0.894}\makecell{54.77\\[-2pt]{\tiny$\pm$11.0}} & \cellcolor[rgb]{0.568,0.783,0.340}\makecell{76.65\\[-2pt]{\tiny$\pm$18.7}} & \cellcolor[rgb]{0.594,0.800,0.372}\makecell{75.52\\[-2pt]{\tiny$\pm$19.8}} \\
\midrule
\multirow{2}{*}{\textit{Average}} & base & \cellcolor[rgb]{0.460,0.705,0.230}{81.85} & \cellcolor[rgb]{0.717,0.880,0.520}{70.06} & \cellcolor[rgb]{0.421,0.673,0.206}{83.98} & \cellcolor[rgb]{0.429,0.679,0.211}{83.57} \\
 & PT & \cellcolor[rgb]{0.533,0.760,0.297}{78.32} & \cellcolor[rgb]{0.725,0.884,0.531}{69.60} & \cellcolor[rgb]{0.483,0.724,0.245}{80.61} & \cellcolor[rgb]{0.490,0.731,0.250}{80.13} \\
\bottomrule
\end{tabular}
\caption{Average model accuracy by evaluation method and model variant. Results are reported for the largest model in each model family, a detailed breakdown of all model scores is provided in Appendix~\ref{app:model_results}. Error range denotes standard deviation.}
\label{tab:model_accuracy}
\end{table}

\section{Experiments Results}
We present our results in six parts. We first report general performance (\S\ref{sec:general_performance}) and direct evaluation accuracy (\S\ref{sec:behavioral_accuracy}), followed by an analysis of differences between evaluation metrics (\S\ref{sec:alignment_tax}--\S\ref{sec:language_prompting}), and item-level predictors of the alignment tax (\S\ref{sec:item_level}). 

\subsection{General Model Performance}\label{sec:general_performance}
We report the general model performance of the largest model in each model family in Table~\ref{tab:model_accuracy}, across the four evaluation methods and split out by base/PT version.
These results demonstrate that the assessment of linguistic ability is strongly driven by evaluation method and model variant.
The base Gemma~4 model obtains the highest score overall, with the \textit{native prompt} method achieving 88.36\% accuracy.
The combination of native prompt method with base model obtains the highest accuracy on average across models as well, demonstrating that conditioning the model on an instruction in the language we evaluate on improves its grammaticality judgment.

Averaged across models, there is substantial divergence between the three discriminative methods and the meta evaluation method, with the discriminative methods ranging from 78.32\% (\textit{direct}, PT) to 83.98\% (\textit{native prompt}, base), and the meta method at 69.6\% (PT) and 70.06\% (base).
However, the variance across models within a specific method is substantially higher for the meta method, with Gemma~4 obtaining 84.38\% and Llama~3 only 54.51\%.
In the next sections we will explore in more detail which model and data properties drive these differences.

\begin{table}[t]
\centering
\scriptsize
\begin{tabular}{@{}ll>{\centering\arraybackslash}p{0.8cm}>{\centering\arraybackslash}p{0.8cm}>{\centering\arraybackslash}p{0.8cm}>{\centering\arraybackslash}p{0.8cm}}
& & \multicolumn{4}{c}{\textit{Resource Tier}} \\ \arrayrulecolor{lightgray}\cmidrule(lr){3-6}\arrayrulecolor{black}
 & & \textbf{Low Zero} & \textbf{Mid} & \textbf{High} & \textbf{Super High} \\
\midrule
\multirow{2}{*}{Gemma~3 (27B)} & base & \cellcolor[rgb]{0.782,0.908,0.622}\makecell{66.72\\[-2pt]{\tiny$\pm$22.1}} & \cellcolor[rgb]{0.367,0.627,0.171}\makecell{86.60\\[-2pt]{\tiny$\pm$9.5}} & \cellcolor[rgb]{0.200,0.449,0.108}\makecell{96.77\\[-2pt]{\tiny$\pm$2.3}} & \cellcolor[rgb]{0.170,0.413,0.102}\makecell{98.60\\[-2pt]{\tiny$\pm$0.6}} \\
 & PT & \cellcolor[rgb]{0.874,0.948,0.770}\makecell{61.70\\[-2pt]{\tiny$\pm$19.5}} & \cellcolor[rgb]{0.475,0.718,0.240}\makecell{81.20\\[-2pt]{\tiny$\pm$10.6}} & \cellcolor[rgb]{0.264,0.527,0.121}\makecell{92.49\\[-2pt]{\tiny$\pm$4.3}} & \cellcolor[rgb]{0.223,0.477,0.113}\makecell{95.15\\[-2pt]{\tiny$\pm$1.6}} \\
\arrayrulecolor{lightgray}\cmidrule(lr){1-6}\arrayrulecolor{black}
\multirow{2}{*}{Gemma~4 (31B)} & base & \cellcolor[rgb]{0.746,0.893,0.565}\makecell{68.61\\[-2pt]{\tiny$\pm$22.4}} & \cellcolor[rgb]{0.337,0.602,0.152}\makecell{88.22\\[-2pt]{\tiny$\pm$8.9}} & \cellcolor[rgb]{0.182,0.428,0.104}\makecell{97.77\\[-2pt]{\tiny$\pm$2.0}} & \cellcolor[rgb]{0.165,0.406,0.100}\makecell{98.86\\[-2pt]{\tiny$\pm$0.7}} \\
 & PT & \cellcolor[rgb]{0.881,0.952,0.782}\makecell{61.08\\[-2pt]{\tiny$\pm$18.2}} & \cellcolor[rgb]{0.559,0.777,0.329}\makecell{76.96\\[-2pt]{\tiny$\pm$10.7}} & \cellcolor[rgb]{0.360,0.621,0.166}\makecell{86.95\\[-2pt]{\tiny$\pm$6.1}} & \cellcolor[rgb]{0.276,0.541,0.124}\makecell{91.79\\[-2pt]{\tiny$\pm$2.6}} \\
\arrayrulecolor{lightgray}\cmidrule(lr){1-6}\arrayrulecolor{black}
\multirow{2}{*}{Llama~3 (8B)} & base & \cellcolor[rgb]{0.810,0.921,0.668}\makecell{64.96\\[-2pt]{\tiny$\pm$20.5}} & \cellcolor[rgb]{0.437,0.686,0.216}\makecell{82.89\\[-2pt]{\tiny$\pm$10.7}} & \cellcolor[rgb]{0.270,0.534,0.123}\makecell{91.91\\[-2pt]{\tiny$\pm$5.4}} & \cellcolor[rgb]{0.194,0.442,0.107}\makecell{97.11\\[-2pt]{\tiny$\pm$1.7}} \\
 & PT & \cellcolor[rgb]{0.838,0.933,0.713}\makecell{63.42\\[-2pt]{\tiny$\pm$19.9}} & \cellcolor[rgb]{0.467,0.711,0.235}\makecell{81.64\\[-2pt]{\tiny$\pm$11.0}} & \cellcolor[rgb]{0.299,0.569,0.129}\makecell{89.88\\[-2pt]{\tiny$\pm$6.1}} & \cellcolor[rgb]{0.223,0.477,0.113}\makecell{95.31\\[-2pt]{\tiny$\pm$2.5}} \\
\arrayrulecolor{lightgray}\cmidrule(lr){1-6}\arrayrulecolor{black}
\multirow{2}{*}{Qwen~3 (30B)} & base & \cellcolor[rgb]{0.845,0.936,0.725}\makecell{63.26\\[-2pt]{\tiny$\pm$20.5}} & \cellcolor[rgb]{0.475,0.718,0.240}\makecell{81.06\\[-2pt]{\tiny$\pm$10.7}} & \cellcolor[rgb]{0.252,0.512,0.119}\makecell{93.15\\[-2pt]{\tiny$\pm$4.6}} & \cellcolor[rgb]{0.194,0.442,0.107}\makecell{97.03\\[-2pt]{\tiny$\pm$1.7}} \\
 & PT & \cellcolor[rgb]{0.874,0.948,0.770}\makecell{61.58\\[-2pt]{\tiny$\pm$19.4}} & \cellcolor[rgb]{0.542,0.766,0.308}\makecell{78.01\\[-2pt]{\tiny$\pm$11.2}} & \cellcolor[rgb]{0.287,0.555,0.126}\makecell{90.85\\[-2pt]{\tiny$\pm$5.8}} & \cellcolor[rgb]{0.229,0.484,0.114}\makecell{94.65\\[-2pt]{\tiny$\pm$2.0}} \\
\arrayrulecolor{lightgray}\cmidrule(lr){1-6}\arrayrulecolor{black}
\multirow{2}{*}{Qwen~3.5 (35B)} & base & \cellcolor[rgb]{0.789,0.912,0.634}\makecell{66.25\\[-2pt]{\tiny$\pm$21.5}} & \cellcolor[rgb]{0.367,0.627,0.171}\makecell{86.36\\[-2pt]{\tiny$\pm$10.2}} & \cellcolor[rgb]{0.217,0.470,0.112}\makecell{95.56\\[-2pt]{\tiny$\pm$2.8}} & \cellcolor[rgb]{0.182,0.428,0.104}\makecell{97.91\\[-2pt]{\tiny$\pm$1.0}} \\
 & PT & \cellcolor[rgb]{0.803,0.918,0.656}\makecell{65.62\\[-2pt]{\tiny$\pm$20.7}} & \cellcolor[rgb]{0.398,0.653,0.191}\makecell{84.98\\[-2pt]{\tiny$\pm$9.5}} & \cellcolor[rgb]{0.235,0.491,0.115}\makecell{94.36\\[-2pt]{\tiny$\pm$3.5}} & \cellcolor[rgb]{0.194,0.442,0.107}\makecell{96.96\\[-2pt]{\tiny$\pm$1.5}} \\
\arrayrulecolor{lightgray}\cmidrule(lr){1-6}\arrayrulecolor{black}
\multirow{2}{*}{TinyAya (3.35B)} & base & \cellcolor[rgb]{0.867,0.945,0.759}\makecell{62.07\\[-2pt]{\tiny$\pm$20.0}} & \cellcolor[rgb]{0.630,0.823,0.414}\makecell{74.14\\[-2pt]{\tiny$\pm$11.2}} & \cellcolor[rgb]{0.337,0.602,0.152}\makecell{88.01\\[-2pt]{\tiny$\pm$11.7}} & \cellcolor[rgb]{0.188,0.435,0.105}\makecell{97.60\\[-2pt]{\tiny$\pm$1.3}} \\
 & PT & \cellcolor[rgb]{0.867,0.945,0.759}\makecell{61.81\\[-2pt]{\tiny$\pm$18.9}} & \cellcolor[rgb]{0.691,0.862,0.488}\makecell{71.19\\[-2pt]{\tiny$\pm$13.1}} & \cellcolor[rgb]{0.398,0.653,0.191}\makecell{85.11\\[-2pt]{\tiny$\pm$11.2}} & \cellcolor[rgb]{0.252,0.512,0.119}\makecell{93.19\\[-2pt]{\tiny$\pm$2.7}} \\
\midrule
\multirow{2}{*}{\textit{Average}} & base & \cellcolor[rgb]{0.803,0.918,0.656}{65.31} & \cellcolor[rgb]{0.429,0.679,0.211}{83.21} & \cellcolor[rgb]{0.241,0.498,0.116}{93.86} & \cellcolor[rgb]{0.182,0.428,0.104}{97.85} \\
 & PT & \cellcolor[rgb]{0.852,0.939,0.736}{62.53} & \cellcolor[rgb]{0.516,0.749,0.276}{79.00} & \cellcolor[rgb]{0.299,0.569,0.129}{89.94} & \cellcolor[rgb]{0.235,0.491,0.115}{94.51} \\
\bottomrule
\end{tabular}
\caption{Average model accuracy by resource tier and model variant, for the \textit{direct} evaluation method.}
\label{tab:lang_tier_accuracy}
\end{table}

\subsection{Direct Evaluation}\label{sec:behavioral_accuracy}
Table~\ref{tab:lang_tier_accuracy} reports mean Direct accuracy for the largest model in each family, averaged equally across languages within each resource tier. Across all model families, accuracy increases monotonically with resource availability. Base models consistently outperform their PT counterparts across all tiers, with the alignment tax more pronounced in lower-resource tiers. Super High Resource languages yield substantially higher accuracy than Low/Zero Resource languages, with gaps of approximately 30--35 percentage points, reflecting the strong influence of pretraining data availability on grammatical competence. All results for individual model sizes and languages can be found in Appendix~\ref{sec:appendix_families}.

Among the six families, Gemma~3 (27B) and Gemma~4 (31B) achieve the highest accuracy across resource tiers, with base models reaching 98.60\% and 98.86\% respectively on Super High languages. 
TinyAya shows competitive performance relative to its size of 3.35B parameters, yet despite its claims about more equal language support, it remains heavily skewed towards high resource languages.
A more detailed comparison of TinyAya against similarly sized models is provided in Appendix~\ref{sec:tiny_aya}.
For Low/Zero Resource languages, all models cluster in the range of 60--69\%, approaching but not reaching the 50\% chance baseline. 


\subsection{The Linguistic Alignment Tax}\label{sec:alignment_tax}
\begin{figure}[t]
    \centering
    \includegraphics[width=\columnwidth]{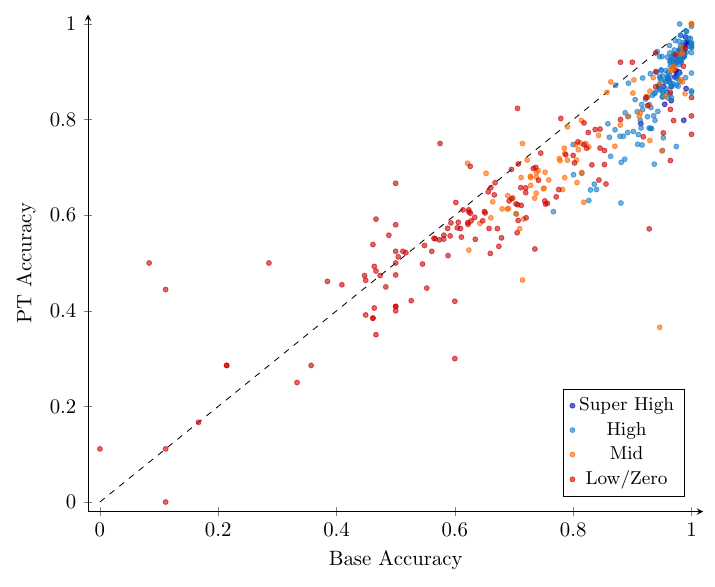}
    \caption{Base vs.\ PT accuracy for
    Gemma~3 across 101 languages. Each point is one (language, model size)
    combination; color indicates resource tier. The dashed diagonal marks
    parity; points below the diagonal indicate linguistic alignment tax.}
    \label{fig:pt_vs_it_gemma3}
\end{figure}

Figure~\ref{fig:pt_vs_it_gemma3} plots base accuracy against
post-trained accuracy for each $\langle$language, model size$\rangle$ combination
in the Gemma~3 family. We focus on Gemma~3 for scaling analyses
throughout \S\ref{sec:alignment_tax}--\ref{sec:inarticulate_gap}, as
it offers the broadest parameter ladder (1B--27B) among the evaluated
families; cross-family comparisons are provided at the end of each
subsection. Two patterns emerge: base and PT accuracy are strongly
correlated, but the majority of points fall below the diagonal,
indicating that post-training consistently \textit{degrades} grammatical
judgment. We term this degradation the \textbf{linguistic alignment
tax}: $\Delta = \mathrm{Acc}_{\mathrm{base}} - \mathrm{Acc}_{\mathrm{PT}}$.

However, raw tax alone can be misleading. For Gemma~3 (27B), Hebrew loses 7.9 percentage points after post-training against 3.0 for English, but in proportional terms the gap is wider still: English (base: 99.2\%) retains 93.9\% of its above-chance knowledge, while Hebrew (base: 92.2\%) retains only 81.2\%. We therefore define the \textit{normalized linguistic
alignment tax}:
\begin{equation}
    \text{Normalized Tax} = 
    \frac{\mathrm{Acc}_{\mathrm{base}} - \mathrm{Acc}_{\mathrm{PT}}}
         {\mathrm{Acc}_{\mathrm{base}} - 0.5}
\end{equation}
Languages where $\mathrm{Acc}_{\mathrm{base}} \leq 0.5$ are excluded
from normalization. In the remainder of this section, ``tax'' refers to the normalized tax. 

\begin{figure}[t]
    \centering
\resizebox{\columnwidth}{!}{\begin{tikzpicture}
\begin{axis}[
    width=12cm, height=6.5cm,
    xlabel={Model Size},
    ylabel={Normalized Tax (\%)},
    xmin=0.5, xmax=4.5,
    ymin=-15, ymax=60,
    xtick={1,2,3,4},
    xticklabels={1B, 4B, 12B, 27B},
    legend pos=north east,
    legend style={font=\small},
    axis lines=left,
    clip=false,
]

\addplot[black!30, dashed, thin, forget plot] coordinates {(0.5,0) (4.5,0)};

\addplot[name path=sh_hi, draw=none, forget plot]
    table[col sep=space, x=size, y=hi] {figures/data/tax_by_tier_gemma3_Super_High.dat};
\addplot[name path=sh_lo, draw=none, forget plot]
    table[col sep=space, x=size, y=lo] {figures/data/tax_by_tier_gemma3_Super_High.dat};
\addplot[blue!70!black, fill opacity=0.15, draw=none, forget plot]
    fill between[of=sh_hi and sh_lo];

\addplot[name path=h_hi, draw=none, forget plot]
    table[col sep=space, x=size, y=hi] {figures/data/tax_by_tier_gemma3_High.dat};
\addplot[name path=h_lo, draw=none, forget plot]
    table[col sep=space, x=size, y=lo] {figures/data/tax_by_tier_gemma3_High.dat};
\addplot[cyan!60!blue, fill opacity=0.15, draw=none, forget plot]
    fill between[of=h_hi and h_lo];

\addplot[name path=m_hi, draw=none, forget plot]
    table[col sep=space, x=size, y=hi] {figures/data/tax_by_tier_gemma3_Mid.dat};
\addplot[name path=m_lo, draw=none, forget plot]
    table[col sep=space, x=size, y=lo] {figures/data/tax_by_tier_gemma3_Mid.dat};
\addplot[orange!80!red, fill opacity=0.15, draw=none, forget plot]
    fill between[of=m_hi and m_lo];

\addplot[name path=lz_hi, draw=none, forget plot]
    table[col sep=space, x=size, y=hi] {figures/data/tax_by_tier_gemma3_Low_Zero.dat};
\addplot[name path=lz_lo, draw=none, forget plot]
    table[col sep=space, x=size, y=lo] {figures/data/tax_by_tier_gemma3_Low_Zero.dat};
\addplot[red!80!black, fill opacity=0.15, draw=none, forget plot]
    fill between[of=lz_hi and lz_lo];

\addplot[blue!70!black, mark=*, mark size=2pt, thick]
    table[col sep=space, x=size, y=mean_tax] {figures/data/tax_by_tier_gemma3_Super_High.dat};
\addplot[cyan!60!blue, mark=*, mark size=2pt, thick]
    table[col sep=space, x=size, y=mean_tax] {figures/data/tax_by_tier_gemma3_High.dat};
\addplot[orange!80!red, mark=*, mark size=2pt, thick]
    table[col sep=space, x=size, y=mean_tax] {figures/data/tax_by_tier_gemma3_Mid.dat};
\addplot[red!80!black, mark=*, mark size=2pt, thick]
    table[col sep=space, x=size, y=mean_tax] {figures/data/tax_by_tier_gemma3_Low_Zero.dat};

\legend{Super High, High, Mid, Low/Zero}

\end{axis}
\end{tikzpicture}}
    \caption{Normalized linguistic alignment tax for Gemma~3 by resource tier and model size. Shaded bands are 95\% bootstrap confidence intervals over the languages in each tier (10{,}000 resamples). 12 languages where $\mathrm{Acc}_{\mathrm{base}} \leq 0.5$ have been excluded from all sizes.}
    \label{fig:tax_by_tier_gemma3}
\end{figure}
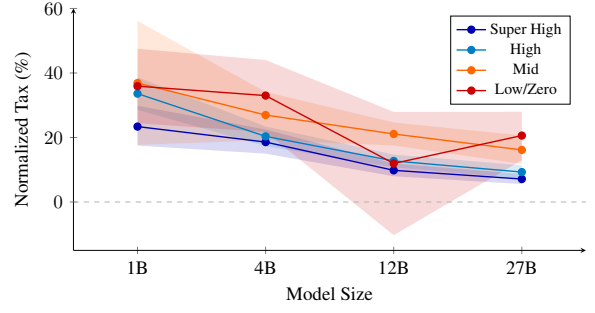

\paragraph{Tax by Resource Tier}
Figure~\ref{fig:tax_by_tier_gemma3} reports the normalized tax for Gemma~3 across model scales and resource tiers. Super High and High Resource languages respond well to scaling: their taxes drop from 23.4\% and 33.6\% at 1B to 7.1\% and 9.3\% at 27B, respectively.
Mid-Resource languages follow a similar but slower trajectory (36.9\% $\rightarrow$ 16.1\%), with diminishing returns beyond 12B.
In the Low/Zero tier, tax drops from 35.9\% at 1B to 12.0\% at 12B and stands at 20.6\% at 27B; this rebound is within bootstrap uncertainty (paired 27B$-$12B difference $+8.6$pp, 95\% CI $[-5.4, +27.3]$, 11 of 24 languages increasing), and the 27B figure holds at 17.2--20.6\% when the tier's smallest test sets are dropped (Appendix~\ref{sec:appendix_smalln}).
Scaling reduces the tax unevenly: low-resource languages keep the highest tax even at the largest size.

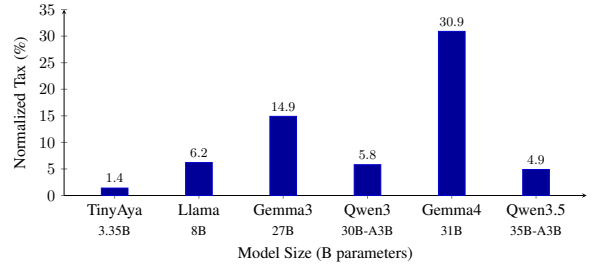
\begin{figure}[t]
    \centering
\resizebox{\columnwidth}{!}{\begin{tikzpicture}
\begin{axis}[
    width=14cm, height=6cm, ytick={0,5,10,15,20,25,30,35},
    ybar,
    xlabel={Model Size (B parameters)},
    ylabel={Normalized Tax (\%)},
    ymin=0, ymax=35,
    symbolic x coords={3.35, 8, 27, 30, 31, 35},
    xtick=data,
    xticklabels={
        TinyAya\\{\footnotesize 3.35B},
        Llama\\{\footnotesize 8B},
        Gemma3\\{\footnotesize 27B},
        Qwen3\\{\footnotesize 30B-A3B},
        Gemma4\\{\footnotesize 31B},
        Qwen3.5\\{\footnotesize 35B-A3B}
    },
    x tick label style={align=center},
    bar width=18pt,
    nodes near coords,
    nodes near coords align={vertical},
    every node near coord/.append style={font=\small},
    axis lines=left,
    clip=false,
    enlarge x limits=0.12,
]

\addplot[fill=blue!60!black, draw=blue!80!black]
    coordinates {
        (3.35, 1.4)
        (8, 6.2)
        (27, 14.9)
        (30, 5.8)
        (31, 30.9)
        (35, 4.9)
    };

\end{axis}
\end{tikzpicture}}
    \caption{Normalized linguistic alignment tax for the largest model in each family, measured using the Direct method, and averaged equally across languages with above-chance base accuracy (91--93 of 101, varying by model)}
    \label{fig:norm_tax_cross_family}
\end{figure}

\paragraph{Cross-Family Comparison}
Figure~\ref{fig:norm_tax_cross_family} reports the normalized tax for
the largest model in each family. The ranking reveals that alignment tax is not driven by model size: TinyAya (3.35B) exhibits the lowest tax (1.4\%), while Gemma~4-31B shows the highest (30.9\%).
Since base accuracies are comparable across families (76.7\%--85.6\%), the divergence does not stem from differences in base competence, indicating that alignment tax is a `model-family-specific' property rather than a universal cost of post-training. Because neither post-training procedures nor pre-training data are disclosed for these models, we cannot isolate which family-level factor is responsible.

\subsection{The Inarticulate Gap}
\label{sec:inarticulate_gap}
We define the \textbf{inarticulate gap} as the difference between a post-trained model's `implicit' and `explicit' grammatical knowledge: $\mathrm{Gap} = \mathrm{Acc}_{\mathrm{Direct}}(\text{PT}) - \mathrm{Acc}_{\mathrm{Meta}}(\text{PT})$.
A positive gap indicates that the model retains grammatical knowledge it cannot articulate through the chat interface. 
As in \S\ref{sec:alignment_tax}, we use Gemma~3 for within-family scaling analyses; cross-family comparisons follow at the end of this subsection. 
This Direct-vs-Meta discrepancy parallels the paradigm used by \citet{song2025languagemodelsfailintrospect} to study introspection, though our focus is on quantifying the knowledge gap rather than testing privileged self-access.

\paragraph{Base Models as a Format Mismatch Baseline}
 Base models also exhibit a positive gap across all resource tiers, indicating that the gap is not solely an artifact of post-training. For the Gemma~3 family, the gap in Super High languages narrows from 0.473 at 1B to 0.070 at 27B, suggesting that larger pre-trained models are less affected by the format mismatch between Direct probability measurements and Meta prompting. We treat the base gap as a structural baseline against which PT convergence is measured (Table~\ref{tab:base_pt_gap}).

\begin{figure}[t]
    \centering
    \includegraphics[width=\columnwidth]{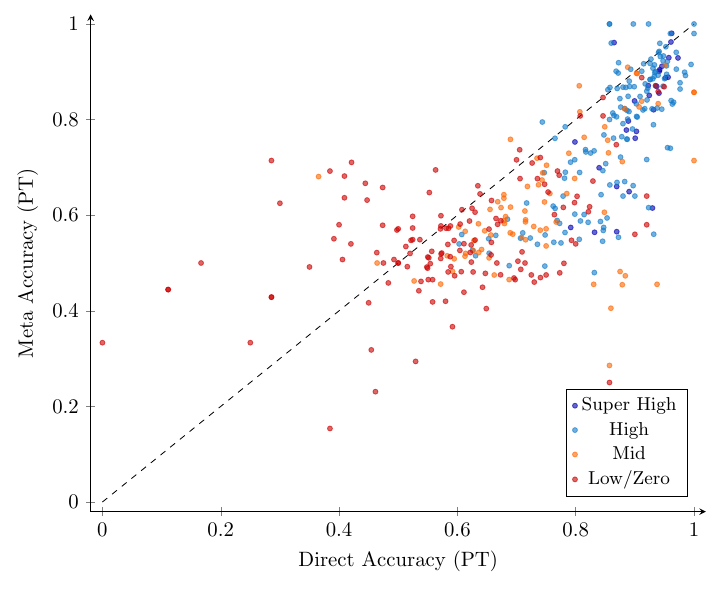}
    \caption{Direct vs.\ Meta accuracy for Gemma~3 under the PT
    setting. Each point is one $\langle$language, model size$\rangle$
    combination; color indicates resource tier. The dashed diagonal
    marks parity; points above the diagonal indicate a positive
    inarticulate gap.}
    \label{fig:inarticulate_gemma3}
\end{figure}

\begin{figure}[t]
    \centering
    \resizebox{\columnwidth}{!}{\begin{tikzpicture}
\begin{axis}[
    width=12cm, height=6.5cm, ytick={0,0.05,0.1,0.15,0.2,0.25,0.3},
    xlabel={Model Size},
    ylabel={Inarticulate Gap},
    xmin=0.5, xmax=4.5,
    ymin=-0.05, ymax=0.30,
    xtick={1,2,3,4},
    xticklabels={1B, 4B, 12B, 27B},
    legend pos=north east,
    legend style={font=\small},
    axis lines=left,
    clip=false,
    yticklabel style={/pgf/number format/fixed},
    scaled y ticks=false,
]

\addplot[black!30, dashed, thin, forget plot] coordinates {(0.5,0) (4.5,0)};

\addplot[name path=sh_hi, draw=none, forget plot]
    table[col sep=space, x=size, y=hi] {figures/data/gap_by_tier_gemma3_Super_High.dat};
\addplot[name path=sh_lo, draw=none, forget plot]
    table[col sep=space, x=size, y=lo] {figures/data/gap_by_tier_gemma3_Super_High.dat};
\addplot[blue!70!black, fill opacity=0.15, draw=none, forget plot]
    fill between[of=sh_hi and sh_lo];

\addplot[name path=h_hi, draw=none, forget plot]
    table[col sep=space, x=size, y=hi] {figures/data/gap_by_tier_gemma3_High.dat};
\addplot[name path=h_lo, draw=none, forget plot]
    table[col sep=space, x=size, y=lo] {figures/data/gap_by_tier_gemma3_High.dat};
\addplot[cyan!60!blue, fill opacity=0.15, draw=none, forget plot]
    fill between[of=h_hi and h_lo];

\addplot[name path=m_hi, draw=none, forget plot]
    table[col sep=space, x=size, y=hi] {figures/data/gap_by_tier_gemma3_Mid.dat};
\addplot[name path=m_lo, draw=none, forget plot]
    table[col sep=space, x=size, y=lo] {figures/data/gap_by_tier_gemma3_Mid.dat};
\addplot[orange!80!red, fill opacity=0.15, draw=none, forget plot]
    fill between[of=m_hi and m_lo];

\addplot[name path=lz_hi, draw=none, forget plot]
    table[col sep=space, x=size, y=hi] {figures/data/gap_by_tier_gemma3_Low_Zero.dat};
\addplot[name path=lz_lo, draw=none, forget plot]
    table[col sep=space, x=size, y=lo] {figures/data/gap_by_tier_gemma3_Low_Zero.dat};
\addplot[red!80!black, fill opacity=0.15, draw=none, forget plot]
    fill between[of=lz_hi and lz_lo];

\addplot[blue!70!black, mark=*, mark size=2pt, thick]
    table[col sep=space, x=size, y=mean_gap] {figures/data/gap_by_tier_gemma3_Super_High.dat};
\addplot[cyan!60!blue, mark=*, mark size=2pt, thick]
    table[col sep=space, x=size, y=mean_gap] {figures/data/gap_by_tier_gemma3_High.dat};
\addplot[orange!80!red, mark=*, mark size=2pt, thick]
    table[col sep=space, x=size, y=mean_gap] {figures/data/gap_by_tier_gemma3_Mid.dat};
\addplot[red!80!black, mark=*, mark size=2pt, thick]
    table[col sep=space, x=size, y=mean_gap] {figures/data/gap_by_tier_gemma3_Low_Zero.dat};

\legend{Super High, High, Mid, Low/Zero}

\end{axis}
\end{tikzpicture}}
    \caption{Mean inarticulate gap for Gemma~3 by resource tier
    and model size. Gap is computed as
    $\mathrm{Acc}_{\mathrm{Direct}}(\text{PT}) -
    \mathrm{Acc}_{\mathrm{Meta}}(\text{PT})$,
    averaged equally across languages within each tier.
    Shaded bands are 95\% bootstrap confidence intervals over the
    languages in each tier (10{,}000 resamples).}
    \label{fig:gap_by_tier_gemma3}
\end{figure}
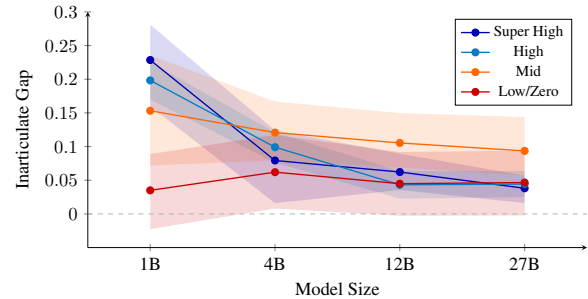

\paragraph{PT Models: Hidden Competence Persists}
Figure~\ref{fig:inarticulate_gemma3} plots Direct vs.\ Meta accuracy for all $\langle$language, model size$\rangle$ pairs under the PT setting, and Figure~\ref{fig:gap_by_tier_gemma3} reports the mean gap by tier and scale. 
The majority of points lie above the diagonal---89.3\% of Super High, 86.8\% of High, 92.5\% of Mid, and 66.7\% of Low/Zero Resource pairs exhibit Direct accuracy exceeding Meta accuracy---confirming the \textit{hidden competence} hypothesis \citep{song2025languagemodelsfailintrospect}. 
Scaling reduces the gap, but unevenly: Super High and High Resource languages improve substantially (Super High: 0.228 $\rightarrow$ 0.038; High: 0.198 $\rightarrow$ 0.044), while Low/Zero Resource languages remain flat (0.035 $\rightarrow$ 0.047), not because scaling fails but because Direct accuracy in this tier is already close to chance, leaving little room for a gap to form. 
The inarticulate gap is therefore primarily a high-resource phenomenon.

\begin{figure}[t]
    \centering
    \resizebox{\columnwidth}{!}{\begin{tikzpicture}
\begin{axis}[
    width=14cm, height=6cm, ytick={-0.1,-0.05,0,0.05,0.1,0.15,0.2,0.25},
    ybar,
    xlabel={Model Size (B parameters)},
    ylabel={Inarticulate Gap},
    ymin=-0.10, ymax=0.25,
    scaled y ticks=false,
    yticklabel style={/pgf/number format/fixed, /pgf/number format/precision=3},
    symbolic x coords={3.35, 8, 27, 30, 31, 35},
    xtick=data,
    xticklabels={
        TinyAya\\{\footnotesize 3.35B},
        Llama\\{\footnotesize 8B},
        Gemma3\\{\footnotesize 27B},
        Qwen3\\{\footnotesize 30B-A3B},
        Gemma4\\{\footnotesize 31B},
        Qwen3.5\\{\footnotesize 35B-A3B}
    },
    x tick label style={align=center},
    bar width=18pt,
    nodes near coords,
    nodes near coords align={vertical},
    nodes near coords style={
        font=\small,
        /pgf/number format/fixed,
        /pgf/number format/fixed zerofill,
        /pgf/number format/precision=3,
    },
    axis lines=left,
    clip=false,
    enlarge x limits=0.12,
]
\addplot[fill=blue!60!black, draw=blue!80!black]
    coordinates {
        (3.35, 0.198)
        (8, 0.219)
        (27, 0.054)
        (30, 0.095)
        (31, -0.062)
        (35, 0.020)
    };
\draw[black, thick] ({rel axis cs:0,0} |- {axis cs:3.35,0}) -- ({rel axis cs:1,0} |- {axis cs:3.35,0});
\end{axis}
\end{tikzpicture}}
    \caption{Mean inarticulate gap for the largest model in each
    family, computed as $\mathrm{Acc}_{\mathrm{Direct}}(\text{PT}) -
    \mathrm{Acc}_{\mathrm{Meta}}(\text{PT})$, averaged equally
    across 101 languages. Models are ordered by total parameter
    count.}
    \label{fig:inarticulate_cross_family}
\end{figure}
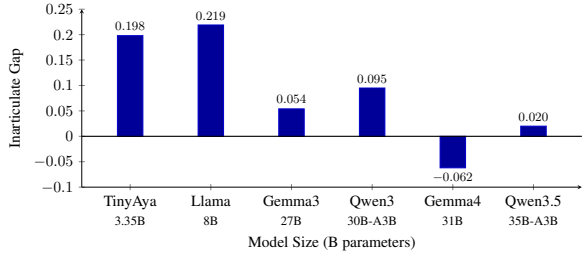

\paragraph{Cross-Family Comparison}
Figure~\ref{fig:inarticulate_cross_family} reports the mean gap for the largest model in each family. 
Llama~3 (0.219) and TinyAya (0.198) exhibit the largest gaps, while Gemma~4 shows a \textit{negative} gap ($-0.062$), indicating that its Meta accuracy exceeds its Direct accuracy. 
As with the alignment tax, base accuracies are comparable across families, so this divergence does not stem from differences in base competence. 
The mean pairwise Spearman correlation of language-level gaps across all six families ($\rho = 0.412$) is substantially higher than the same analysis for alignment tax ($\rho = 0.206$), indicating that the gap is more strongly driven by language-level properties, while alignment tax is more sensitive to family-level factors.

\paragraph{Gap, Alignment Tax, and Pre-training}
Across languages the gap and the alignment tax move in opposite directions: in four of the six families they are moderately negatively correlated ($\rho$ from $-0.26$ to $-0.58$, all $p<0.05$), while in Qwen~3 and Qwen~3.5 there is no relationship.
The languages that lose the most to post-training are therefore not the languages where the most grammatical knowledge goes unarticulated.
Furthermore, the base gap is a reliable predictor of the PT gap ($\rho = 0.603$, $p < 0.001$, consistent across all families), suggesting that `inarticulation' arises during pre-training: languages where the base model struggles to express grammatical knowledge in chat format also exhibit larger gaps after post-training.

\begin{figure*}[t]
    \centering
    \includegraphics[width=5.5in]{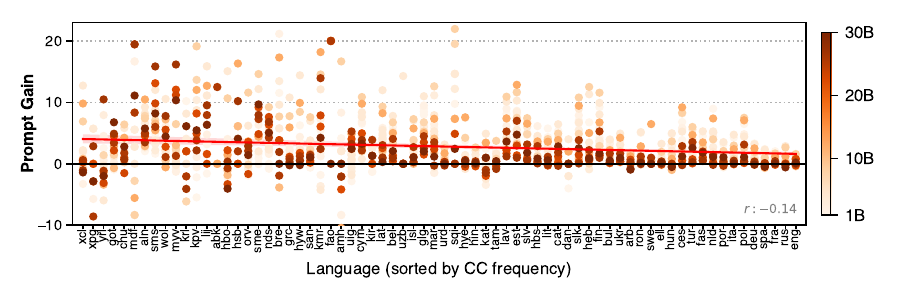}
    \caption{Prompt Gain set out against language frequency across all evaluated models, which are colored by size in billions of parameters.}
    \label{fig:prompt_gain}
\end{figure*}

\subsection{Language-specific Prompting}\label{sec:language_prompting}
To investigate the role of language-specific prompting in eliciting grammatical knowledge, we introduce a complementary metric, the \textbf{Prompt Gain}: $\mathrm{Acc}_{\mathrm{Native}}(\text{PT}) - \mathrm{Acc}_{\mathrm{Direct}}(\text{PT})$, the difference between accuracy under native prompting and under direct evaluation.
While the inarticulate gap captures the failure to articulate latent knowledge under English instructions, the Prompt Gain quantifies how much performance can be improved by providing instructions in the target language itself. 
A model may require a language-specific cue to adapt its probability distribution to the linguistic structure specific to that language.
A positive value then indicates that native-language prompting successfully `activates' linguistic knowledge that remains latent under direct evaluation.

Figure~\ref{fig:prompt_gain} reveals a striking cross-lingual gradient in the Prompt Gain. 
For high-resource languages, the gain approaches zero, indicating that explicit prompting provides minimal benefit over the model's raw probabilistic judgments: the model's base predictions are sufficiently robust and tailored towards these high-resource languages that grammatical ability manifests accurately without language-specific prompting.
For low-resource languages, the advantage is substantially more positive, demonstrating that explicit, language-specific instructions help the model apply grammatical knowledge that is present but weakly expressed in raw probabilities.\footnote{For 13 of the 36 Low/Zero languages the native cue is a substitute rather than a translation; these lower the tier's mean Prompt Gain rather than raise it (Appendix~\ref{sec:appendix_subcue}).} 
This pattern is stronger in smaller models for mid- to high-resource languages: smaller models thus need explicit prompting to reliably access their grammatical knowledge, while larger models can apply it more directly.
This pattern supports \citet{poelman-lhoneux-2025-roles}'s argument against using English as an `evaluation interface': native prompting substantially improves low-resource language performance, indicating that direct probability and English prompting underestimate grammatical competence where language-specific contextualization is absent.

\subsection{Item-level Analysis of the Alignment Tax}\label{sec:item_level}
The alignment tax is an aggregated metric: post-training may lose items the base model solved, but it also gains some, and the aggregate drop reports only the difference. To see the two flows separately, we pair items across the base and post-trained checkpoints of each model by language and item identifier, under the Direct method.
For Gemma~4 (31B), 78.2\% of the 121{,}305 items are solved by both variants and 4.5\% by neither, while 14.5\% are lost (base correct, post-trained incorrect) and 2.9\% are gained (base incorrect, post-trained correct); the difference between the two flip classes, 11.6 percentage points, is exactly the aggregate tax. 
Flips are thus strongly one-directional. 
Losses outnumber gains in all six families, by ratios from 1.4:1 (Qwen~3.5) to 5.0:1 (Gemma~4).

\begin{figure}[t]
    \centering
    \resizebox{\columnwidth}{!}{\begin{tikzpicture}
\begin{axis}[
    width=12cm, height=6.5cm, ytick={0,10,20,30,40,50},
    xlabel={Base-model margin $\log P(S_{cor})-\log P(S_{inc})$},
    ylabel={Items lost after post-training (\%)},
    xmin=0.5, xmax=9.5,
    ymin=0, ymax=50,
    xtick={1,2,3,4,5,6,7,8,9},
    xticklabels={0--1, 1--2, 2--3, 3--4, 4--6, 6--8, 8--12, 12--20, $>$20},
    xticklabel style={font=\small},
    legend pos=north east,
    legend style={font=\small},
    axis lines=left,
    clip=false,
    yticklabel style={/pgf/number format/fixed},
    scaled y ticks=false,
]

\addplot[name path=sh_hi, draw=none, forget plot]
    table[col sep=space, x=bin, y=hi] {figures/data/loss_by_margin_gemma4_Super_High.dat};
\addplot[name path=sh_lo, draw=none, forget plot]
    table[col sep=space, x=bin, y=lo] {figures/data/loss_by_margin_gemma4_Super_High.dat};
\addplot[blue!70!black, fill opacity=0.15, draw=none, forget plot]
    fill between[of=sh_hi and sh_lo];

\addplot[name path=h_hi, draw=none, forget plot]
    table[col sep=space, x=bin, y=hi] {figures/data/loss_by_margin_gemma4_High.dat};
\addplot[name path=h_lo, draw=none, forget plot]
    table[col sep=space, x=bin, y=lo] {figures/data/loss_by_margin_gemma4_High.dat};
\addplot[cyan!60!blue, fill opacity=0.15, draw=none, forget plot]
    fill between[of=h_hi and h_lo];

\addplot[name path=m_hi, draw=none, forget plot]
    table[col sep=space, x=bin, y=hi] {figures/data/loss_by_margin_gemma4_Mid.dat};
\addplot[name path=m_lo, draw=none, forget plot]
    table[col sep=space, x=bin, y=lo] {figures/data/loss_by_margin_gemma4_Mid.dat};
\addplot[orange!80!red, fill opacity=0.15, draw=none, forget plot]
    fill between[of=m_hi and m_lo];

\addplot[name path=lz_hi, draw=none, forget plot]
    table[col sep=space, x=bin, y=hi] {figures/data/loss_by_margin_gemma4_Low_Zero.dat};
\addplot[name path=lz_lo, draw=none, forget plot]
    table[col sep=space, x=bin, y=lo] {figures/data/loss_by_margin_gemma4_Low_Zero.dat};
\addplot[red!80!black, fill opacity=0.15, draw=none, forget plot]
    fill between[of=lz_hi and lz_lo];

\addplot[blue!70!black, mark=*, mark size=2pt, thick]
    table[col sep=space, x=bin, y=rate] {figures/data/loss_by_margin_gemma4_Super_High.dat};
\addplot[cyan!60!blue, mark=*, mark size=2pt, thick]
    table[col sep=space, x=bin, y=rate] {figures/data/loss_by_margin_gemma4_High.dat};
\addplot[orange!80!red, mark=*, mark size=2pt, thick]
    table[col sep=space, x=bin, y=rate] {figures/data/loss_by_margin_gemma4_Mid.dat};
\addplot[red!80!black, mark=*, mark size=2pt, thick]
    table[col sep=space, x=bin, y=rate] {figures/data/loss_by_margin_gemma4_Low_Zero.dat};

\legend{Super High, High, Mid, Low/Zero}

\end{axis}
\end{tikzpicture}}
    \caption{Share of items solved by the base model that are lost after post-training, as a function of the base-model margin, for Gemma~4 (31B) under Direct evaluation. Bands are 95\% Wilson intervals. Items whose stored margin rounds to zero (0.9\%) fall outside the first bin.}
    \label{fig:loss_by_margin}
\end{figure}
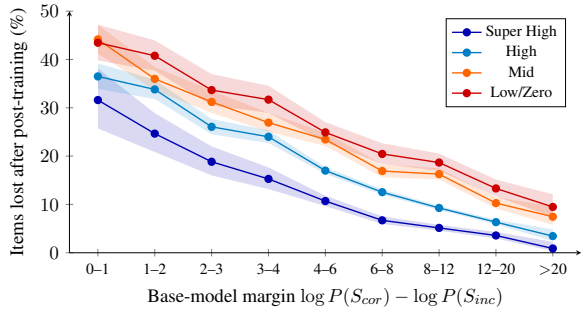

What post-training loses is largely what the base model held weakly. 
Among the items Gemma~4 (31B) solves as a base model, the median margin $\log P(S_{cor})-\log P(S_{inc})$ is 5.0 for items that are subsequently lost and 8.0 for items that are retained; the margin alone ranks lost above retained items with an area under the ROC curve of 0.675, the lowest of the six families, which reach up to 0.821. 
Difficulty alone, however, does not account for the resource gradient. 
Raw loss rates rise from 8.0\% in Super High to 25.4\% in Low/Zero, and a logistic regression of loss on margin, head--verb distance, attractor count, agreement feature and resource tier leaves the tier effect intact: at equal margin the odds of losing an item in Low/Zero exceed those in Super High by a factor of 2.3--4.3 in every family (all $p<0.001$). 
Figure~\ref{fig:loss_by_margin} shows the pattern directly. 

Less frequent agreement feature values are more fragile as well: for Gemma~4 (31B) the raw loss rate is 37.4\% for dual, 21.5\% for plural and 17.1\% for second person, against 11.7\% for third person. 
Two English items lost by this model make the flips concrete; the base model prefers the grammatical form in each pair and the post-trained model the starred one:
\begin{quote}\small
(1) Many details \textbf{remain} / *\textbf{remains} to be worked out, but we will keep you advised.\\
(2) The points system \textbf{acts} / *\textbf{act} as a guide to sort out the best candidates.
\end{quote}
Neither has a number-mismatched noun between the subject and the verb, which is typical: 37 of the 43 English items this model loses have no attractor at all.

\section{Discussion}
Our findings reveal two core properties of multilingual grammatical knowledge in LLMs. First, post-training consistently degrades grammatical competence across languages---the alignment tax---with the degradation amplified in low-resource languages. Second, post-trained models retain grammatical knowledge they cannot express through the chat interface---the inarticulate gap---which is driven by language-level properties rather than family-level factors. Together, these patterns show that linguistic knowledge acquired during pre-training is systematically filtered through post-training, leading to losses that depend on both resource availability and evaluation methods.

A methodological implication of our work is that conclusions about a model's grammatical competence are highly evaluation-dependent. Meta-linguistic prompting yields accuracy ranging from near-chance (Llama~3 at 55\%) to high (Gemma~4 at 84\%) across families with comparable base competence, revealing that it conflates grammatical knowledge with instruction-following ability in a model-family-specific way. 
Future multilingual evaluation should therefore report multiple paradigms rather than treating any single method as definitive.



On the practical side, our native prompt results offer a mitigation: providing evaluation instructions in the target language consistently improves accuracy for low-resource languages, suggesting that {direct} evaluation may actually be \textit{underestimating} low-resource performance.
Our protocol measures immediate grammatical judgment: thinking mode is disabled for the families that support it, so how chain-of-thought reasoning affects a model's ability to articulate grammatical knowledge remains open.
Understanding the mechanisms that trigger linguistic abilities for such languages is an exciting direction for future work, adapting findings from multilingual interpretability in an actionable way to reduce the detrimental effects of language disparity \citep{brinkmann-etal-2025-large, kryvosheieva2026differenttypessyntacticagreement}.

\section*{Limitations}
\paragraph{Benchmark}
All our experiments are conducted on MultiBLiMP, which covers a single class of phenomena: subject-verb and subject-participle agreement in number, person and gender. Our conclusions therefore do not generalize from agreement to grammar more broadly. Agreement captures only a small part of grammaticality, but MultiBLiMP remains the only resource that allows linguistic ability to be measured at this multilingual scale. 
The patterns we report should be read primarily as properties of agreement processing, and may be prone to limitations propagating from MultiBLiMP itself.
Low-resource languages have a smaller set of minimal pairs in MultiBLiMP, and therefore result in a more uncertain estimate of model performance.

\paragraph{Training Data}
Another limitation is about the resource tiers that are from Common Crawl corpus counts, since training data has not been released for any of the evaluated models.
Actual per-language exposure during pre-training may differ substantially across model families, so a family's behavior within a given tier may reflect its own training mixture rather than the tier itself, and cross-family tier-level comparisons should be interpreted with caution.

\paragraph{Post-training Recipe}
Alignment tax varies by an order of magnitude across families at comparable base accuracy, but post-training procedures are not documented for any model we evaluate, so we cannot attribute this variance to any specific aspect of post-training. 
Families also differ in tokenizer segmentation, chat-template formatting, and architecture (dense vs. mixture-of-experts), each of which could contribute. 
Isolating the role of post-training would require applying different recipes to a fixed base model.
A comprehensive study training models from scratch on different training data mixtures would provide a valuable insight into the degree to which linguistic abilities are driven by typological features.

\section*{Acknowledgments}
Jaap Jumelet is supported by NWO grant \texttt{VI.Vidi.221C.009}. 
This work used the Dutch national e-infrastructure with the support of the SURF Cooperative using grant no. \texttt{EINF-13403}.

\bibliography{custom}


\appendix

\onecolumn
\section{Language List}
\label{sec:appendix}

\begin{table*}[h]
\centering
\small
\setlength{\tabcolsep}{2pt}
\begin{tabular}{@{}llr@{\quad}llr@{\quad}llr@{\quad}llr@{}}
\toprule
\multicolumn{3}{c}{\textbf{Super High}} & \multicolumn{3}{c}{\textbf{High}} & \multicolumn{3}{c}{\textbf{Mid}} & \multicolumn{3}{c}{\textbf{Low/Zero}} \\
\cmidrule(lr){1-3} \cmidrule(lr){4-6} \cmidrule(lr){7-9} \cmidrule(lr){10-12}
ISO & Lang & N & ISO & Lang & N & ISO & Lang & N & ISO & Lang & N \\
\midrule
deu & German & 2298 & arb & Arabic & 1215 & amh & Amharic & 112 & abk & Abkhazian & 40 \\
eng & English & 770 & bel & Belarusian & 2570 & bre & Breton & 260 & aln & Gheg Alb.\ & 677 \\
fra & French & 2548 & ben & Bengali & 21 & bua & Buriat & 103 & apu & Apurin\~{a} & 28 \\
ita & Italian & 2999 & bul & Bulgarian & 2458 & cym & Welsh & 1120 & aqz & Akuntsu & 14 \\
pol & Polish & 3272 & cat & Catalan & 2284 & fao & Faroese & 232 & azz & H-P Nahuatl & 207 \\
rus & Russian & 3832 & ces & Czech & 4256 & fro & Old French & 1976 & bho & Bhojpuri & 34 \\
spa & Spanish & 2541 & dan & Danish & 50 & gla & Scot.\ Gaelic & 66 & bor & Bor\^o{}ro & 241 \\
 & &  & ell & Mod.\ Greek & 1096 & gle & Irish & 28 & chu & Church Slav.\ & 4166 \\
 & &  & est & Estonian & 2575 & grc & Anc.\ Greek & 3695 & egy & Egyptian & 22 \\
 & &  & eus & Basque & 273 & guj & Gujarati & 7 & frm & Mid.\ French & 294 \\
 & &  & fas & Persian & 2553 & hsb & Up.\ Sorbian & 186 & got & Gothic & 1579 \\
 & &  & fin & Finnish & 2570 & hyw & W.\ Armenian & 1153 & hbo & Anc.\ Hebrew & 983 \\
 & &  & glg & Galician & 753 & kir & Kyrgyz & 185 & hit & Hittite & 50 \\
 & &  & hbs & Serbo-Croat.\ & 3286 & kmr & N.\ Kurdish & 544 & koi & Komi-Perm.\ & 43 \\
 & &  & heb & Hebrew & 2330 & nds & Low German & 1774 & kpv & Komi-Zyr.\ & 320 \\
 & &  & hin & Hindi & 1447 & orv & Old Russian & 4615 & krl & Karelian & 260 \\
 & &  & hun & Hungarian & 845 & sah & Yakut & 144 & kxh & Karo & 120 \\
 & &  & hye & Armenian & 1415 & san & Sanskrit & 4442 & lij & Ligurian & 254 \\
 & &  & isl & Icelandic & 2801 & sme & N.\ Sami & 2536 & mdf & Moksha & 82 \\
 & &  & kat & Georgian & 204 & uig & Uighur & 758 & myv & Erzya & 464 \\
 & &  & kaz & Kazakh & 173 &  & &  & nhi & T.\ Nahuatl & 38 \\
 & &  & lat & Latin & 3149 &  & &  & olo & Livvi & 190 \\
 & &  & lav & Latvian & 3032 &  & &  & ota & Ott.\ Turkish & 99 \\
 & &  & lit & Lithuanian & 1180 &  & &  & pcm & Nig.\ Pidgin & 26 \\
 & &  & mar & Marathi & 460 &  & &  & quc & K'iche' & 131 \\
 & &  & mkd & Macedonian & 39 &  & &  & sms & Skolt Sami & 263 \\
 & &  & nld & Dutch & 2331 &  & &  & tpn & Tupinamb\'{a} & 9 \\
 & &  & por & Portuguese & 3048 &  & &  & ttc & Tektiteko & 69 \\
 & &  & ron & Romanian & 2056 &  & &  & urb & Urub\'u{}-Kaap.\ & 13 \\
 & &  & slk & Slovak & 4145 &  & &  & vep & Veps & 187 \\
 & &  & slv & Slovenian & 4483 &  & &  & wbp & Warlpiri & 12 \\
 & &  & sqi & Albanian & 243 &  & &  & wol & Wolof & 705 \\
 & &  & swe & Swedish & 201 &  & &  & xcl & Class.\ Armenian & 1623 \\
 & &  & tam & Tamil & 382 &  & &  & xnr & Kangri & 86 \\
 & &  & tur & Turkish & 1742 &  & &  & xpg & Phrygian & 50 \\
 & &  & ukr & Ukrainian & 2744 &  & &  & yrl & Nhengatu & 720 \\
 & &  & urd & Urdu & 550 &  & &  &  & &  \\
 & &  & uzb & Uzbek & 50 &  & &  &  & &  \\
\bottomrule
\end{tabular}
\caption{All 101 languages in the MultiBLiMP benchmark, organized by resource tier. N\,=\,number of minimal pairs. SH\,=\,Super High ($>$10$^{11}$ CC tokens), H\,=\,High (10$^{9}$--10$^{11}$), M\,=\,Mid (10$^{7}$--10$^{9}$), LZ\,=\,Low/Zero ($<$10$^{7}$).}\label{tab:language_list}
\end{table*}

\clearpage
\section{Average Model Accuracy}\label{app:model_results}
Table \ref{tab:model_results_all_models} reports the average model accuracy for each model that we evaluated, for the four different evaluation methods.

\begin{table}[htbp]
\centering
\scriptsize
\begin{tabular}{@{}lll>{\centering\arraybackslash}p{0.8cm}>{\centering\arraybackslash}p{0.8cm}>{\centering\arraybackslash}p{0.8cm}>{\centering\arraybackslash}p{0.8cm}}
& & & \multicolumn{4}{c}{\textit{Evaluation Method}} \\ \arrayrulecolor{lightgray}\cmidrule(lr){4-7}\arrayrulecolor{black}
 & Size & Variant & \textbf{direct} & \textbf{meta} & \textbf{\makecell{native \\ prompt}} & \textbf{\makecell{English \\ prompt}} \\
\midrule
\multirow{8}{*}{Gemma~3} & \multirow{2}{*}{1.0} & PT & \cellcolor[rgb]{0.725,0.884,0.531}67.98 & \cellcolor[rgb]{0.939,0.965,0.900}54.66 & \cellcolor[rgb]{0.682,0.857,0.478}70.52 & \cellcolor[rgb]{0.700,0.868,0.499}69.75 \\
 &  & base & \cellcolor[rgb]{0.559,0.777,0.329}77.34 & \cellcolor[rgb]{0.970,0.963,0.967}49.58 & \cellcolor[rgb]{0.507,0.743,0.266}79.52 & \cellcolor[rgb]{0.516,0.749,0.276}79.05 \\
 & \multirow{2}{*}{4.0} & PT & \cellcolor[rgb]{0.621,0.817,0.403}74.43 & \cellcolor[rgb]{0.803,0.918,0.656}65.55 & \cellcolor[rgb]{0.568,0.783,0.340}75.26 & \cellcolor[rgb]{0.568,0.783,0.340}73.97 \\
 &  & base & \cellcolor[rgb]{0.467,0.711,0.235}81.38 & \cellcolor[rgb]{0.902,0.961,0.816}60.03 & \cellcolor[rgb]{0.437,0.686,0.216}82.84 & \cellcolor[rgb]{0.444,0.692,0.220}82.49 \\
 & \multirow{2}{*}{12.0} & PT & \cellcolor[rgb]{0.524,0.754,0.287}77.91 & \cellcolor[rgb]{0.673,0.851,0.467}72.16 & \cellcolor[rgb]{0.483,0.724,0.245}79.87 & \cellcolor[rgb]{0.490,0.731,0.250}77.53 \\
 &  & base & \cellcolor[rgb]{0.429,0.679,0.211}83.54 & \cellcolor[rgb]{0.796,0.915,0.645}65.89 & \cellcolor[rgb]{0.398,0.653,0.191}84.89 & \cellcolor[rgb]{0.406,0.660,0.196}84.65 \\
 & \multirow{2}{*}{27.0} & PT & \cellcolor[rgb]{0.483,0.724,0.245}79.46 & \cellcolor[rgb]{0.630,0.823,0.414}74.02 & \cellcolor[rgb]{0.452,0.699,0.225}81.63 & \cellcolor[rgb]{0.467,0.711,0.235}79.69 \\
 &  & base & \cellcolor[rgb]{0.413,0.666,0.201}84.17 & \cellcolor[rgb]{0.647,0.834,0.435}73.17 & \cellcolor[rgb]{0.375,0.634,0.176}86.21 & \cellcolor[rgb]{0.383,0.640,0.181}85.77 \\
\arrayrulecolor{lightgray}\cmidrule(lr){1-7}\arrayrulecolor{black}
\multirow{8}{*}{Gemma~4} & \multirow{2}{*}{2.0} & PT & \cellcolor[rgb]{0.665,0.845,0.457}72.28 & \cellcolor[rgb]{0.845,0.936,0.725}63.06 & \cellcolor[rgb]{0.630,0.823,0.414}74.05 & \cellcolor[rgb]{0.638,0.828,0.425}73.76 \\
 &  & base & \cellcolor[rgb]{0.483,0.724,0.245}80.86 & \cellcolor[rgb]{0.967,0.968,0.966}50.15 & \cellcolor[rgb]{0.437,0.686,0.216}82.93 & \cellcolor[rgb]{0.452,0.699,0.225}82.03 \\
 & \multirow{2}{*}{4.0} & PT & \cellcolor[rgb]{0.612,0.811,0.393}74.61 & \cellcolor[rgb]{0.845,0.936,0.725}63.03 & \cellcolor[rgb]{0.551,0.771,0.319}77.42 & \cellcolor[rgb]{0.568,0.783,0.340}76.89 \\
 &  & base & \cellcolor[rgb]{0.444,0.692,0.220}82.63 & \cellcolor[rgb]{0.817,0.924,0.679}64.78 & \cellcolor[rgb]{0.390,0.647,0.186}85.17 & \cellcolor[rgb]{0.421,0.673,0.206}83.90 \\
 & \multirow{2}{*}{26.0} & PT & \cellcolor[rgb]{0.621,0.817,0.403}74.56 & \cellcolor[rgb]{0.542,0.766,0.308}77.75 & \cellcolor[rgb]{0.577,0.788,0.350}76.39 & \cellcolor[rgb]{0.568,0.783,0.340}76.59 \\
 &  & base & \cellcolor[rgb]{0.421,0.673,0.206}83.96 & \cellcolor[rgb]{0.586,0.794,0.361}76.06 & \cellcolor[rgb]{0.360,0.621,0.166}86.86 & \cellcolor[rgb]{0.367,0.627,0.171}86.54 \\
 & \multirow{2}{*}{31.0} & PT & \cellcolor[rgb]{0.586,0.794,0.361}76.09 & \cellcolor[rgb]{0.452,0.699,0.225}82.33 & \cellcolor[rgb]{0.542,0.766,0.308}78.06 & \cellcolor[rgb]{0.498,0.737,0.255}79.77 \\
 &  & base & \cellcolor[rgb]{0.383,0.640,0.181}85.56 & \cellcolor[rgb]{0.406,0.660,0.196}84.38 & \cellcolor[rgb]{0.329,0.595,0.147}88.36 & \cellcolor[rgb]{0.337,0.602,0.152}88.13 \\
\arrayrulecolor{lightgray}\cmidrule(lr){1-7}\arrayrulecolor{black}
\multirow{6}{*}{Llama~3} & \multirow{2}{*}{1.0} & PT & \cellcolor[rgb]{0.682,0.857,0.478}70.96 & \cellcolor[rgb]{0.969,0.967,0.968}49.62 & \cellcolor[rgb]{0.638,0.828,0.425}73.32 & \cellcolor[rgb]{0.647,0.834,0.435}73.19 \\
 &  & base & \cellcolor[rgb]{0.594,0.800,0.372}75.68 & \cellcolor[rgb]{0.970,0.963,0.967}49.46 & \cellcolor[rgb]{0.542,0.766,0.308}77.98 & \cellcolor[rgb]{0.577,0.788,0.350}76.53 \\
 & \multirow{2}{*}{3.0} & PT & \cellcolor[rgb]{0.612,0.811,0.393}74.34 & \cellcolor[rgb]{0.944,0.966,0.912}53.76 & \cellcolor[rgb]{0.551,0.771,0.319}77.23 & \cellcolor[rgb]{0.568,0.783,0.340}76.12 \\
 &  & base & \cellcolor[rgb]{0.524,0.754,0.287}78.59 & \cellcolor[rgb]{0.962,0.968,0.954}50.89 & \cellcolor[rgb]{0.475,0.718,0.240}81.21 & \cellcolor[rgb]{0.490,0.731,0.250}80.26 \\
 & \multirow{2}{*}{8.0} & PT & \cellcolor[rgb]{0.498,0.737,0.255}79.19 & \cellcolor[rgb]{0.920,0.963,0.858}57.31 & \cellcolor[rgb]{0.452,0.699,0.225}82.07 & \cellcolor[rgb]{0.460,0.705,0.230}81.39 \\
 &  & base & \cellcolor[rgb]{0.475,0.718,0.240}80.88 & \cellcolor[rgb]{0.939,0.965,0.900}54.51 & \cellcolor[rgb]{0.429,0.679,0.211}83.24 & \cellcolor[rgb]{0.444,0.692,0.220}82.74 \\
\arrayrulecolor{lightgray}\cmidrule(lr){1-7}\arrayrulecolor{black}
\multirow{12}{*}{Qwen~3} & \multirow{2}{*}{0.6} & PT & \cellcolor[rgb]{0.725,0.884,0.531}70.39 & \cellcolor[rgb]{0.967,0.968,0.966}50.34 & \cellcolor[rgb]{0.673,0.851,0.467}72.32 & \cellcolor[rgb]{0.682,0.857,0.478}71.81 \\
 &  & base & \cellcolor[rgb]{0.647,0.834,0.435}73.16 & \cellcolor[rgb]{0.962,0.968,0.954}50.90 & \cellcolor[rgb]{0.603,0.806,0.382}75.14 & \cellcolor[rgb]{0.630,0.823,0.414}74.19 \\
 & \multirow{2}{*}{1.7} & PT & \cellcolor[rgb]{0.656,0.840,0.446}72.91 & \cellcolor[rgb]{0.925,0.964,0.870}56.47 & \cellcolor[rgb]{0.594,0.800,0.372}74.76 & \cellcolor[rgb]{0.603,0.806,0.382}73.84 \\
 &  & base & \cellcolor[rgb]{0.594,0.800,0.372}75.50 & \cellcolor[rgb]{0.925,0.964,0.870}56.52 & \cellcolor[rgb]{0.542,0.766,0.308}77.78 & \cellcolor[rgb]{0.551,0.771,0.319}77.35 \\
 & \multirow{2}{*}{4.0} & PT & \cellcolor[rgb]{0.621,0.817,0.403}75.34 & \cellcolor[rgb]{0.941,0.965,0.906}54.08 & \cellcolor[rgb]{0.542,0.766,0.308}77.07 & \cellcolor[rgb]{0.568,0.783,0.340}76.71 \\
 &  & base & \cellcolor[rgb]{0.559,0.777,0.329}77.27 & \cellcolor[rgb]{0.824,0.927,0.691}64.43 & \cellcolor[rgb]{0.490,0.731,0.250}80.35 & \cellcolor[rgb]{0.507,0.743,0.266}79.44 \\
 & \multirow{2}{*}{8.0} & PT & \cellcolor[rgb]{0.533,0.760,0.297}78.18 & \cellcolor[rgb]{0.965,0.968,0.960}50.65 & \cellcolor[rgb]{0.490,0.731,0.250}79.79 & \cellcolor[rgb]{0.507,0.743,0.266}79.50 \\
 &  & base & \cellcolor[rgb]{0.673,0.851,0.467}72.22 & \cellcolor[rgb]{0.717,0.880,0.520}70.17 & \cellcolor[rgb]{0.460,0.705,0.230}81.87 & \cellcolor[rgb]{0.460,0.705,0.230}81.80 \\
 & \multirow{2}{*}{14.0} & PT & \cellcolor[rgb]{0.516,0.749,0.276}78.54 & \cellcolor[rgb]{0.673,0.851,0.467}72.09 & \cellcolor[rgb]{0.467,0.711,0.235}80.31 & \cellcolor[rgb]{0.483,0.724,0.245}80.05 \\
 &  & base & \cellcolor[rgb]{0.533,0.760,0.297}78.25 & \cellcolor[rgb]{0.673,0.851,0.467}72.13 & \cellcolor[rgb]{0.437,0.686,0.216}82.86 & \cellcolor[rgb]{0.452,0.699,0.225}82.15 \\
 & \multirow{2}{*}{30.0} & PT & \cellcolor[rgb]{0.507,0.743,0.266}78.14 & \cellcolor[rgb]{0.746,0.893,0.565}68.67 & \cellcolor[rgb]{0.483,0.724,0.245}80.36 & \cellcolor[rgb]{0.490,0.731,0.250}79.80 \\
 &  & base & \cellcolor[rgb]{0.490,0.731,0.250}80.37 & \cellcolor[rgb]{0.691,0.862,0.488}71.40 & \cellcolor[rgb]{0.444,0.692,0.220}82.56 & \cellcolor[rgb]{0.460,0.705,0.230}81.73 \\
\arrayrulecolor{lightgray}\cmidrule(lr){1-7}\arrayrulecolor{black}
\multirow{10}{*}{Qwen~3.5} & \multirow{2}{*}{0.8} & PT & \cellcolor[rgb]{0.612,0.811,0.393}74.31 & \cellcolor[rgb]{0.939,0.965,0.900}54.35 & \cellcolor[rgb]{0.568,0.783,0.340}76.10 & \cellcolor[rgb]{0.594,0.800,0.372}75.88 \\
 &  & base & \cellcolor[rgb]{0.612,0.811,0.393}74.84 & \cellcolor[rgb]{0.925,0.964,0.870}56.64 & \cellcolor[rgb]{0.542,0.766,0.308}77.93 & \cellcolor[rgb]{0.568,0.783,0.340}76.91 \\
 & \multirow{2}{*}{2.0} & PT & \cellcolor[rgb]{0.542,0.766,0.308}77.57 & \cellcolor[rgb]{0.874,0.948,0.770}61.41 & \cellcolor[rgb]{0.498,0.737,0.255}79.23 & \cellcolor[rgb]{0.524,0.754,0.287}78.96 \\
 &  & base & \cellcolor[rgb]{0.524,0.754,0.287}78.89 & \cellcolor[rgb]{0.852,0.939,0.736}62.56 & \cellcolor[rgb]{0.475,0.718,0.240}80.91 & \cellcolor[rgb]{0.490,0.731,0.250}80.14 \\
 & \multirow{2}{*}{4.0} & PT & \cellcolor[rgb]{0.507,0.743,0.266}78.45 & \cellcolor[rgb]{0.760,0.899,0.588}67.72 & \cellcolor[rgb]{0.452,0.699,0.225}81.34 & \cellcolor[rgb]{0.475,0.718,0.240}80.72 \\
 &  & base & \cellcolor[rgb]{0.542,0.766,0.308}77.82 & \cellcolor[rgb]{0.691,0.862,0.488}71.33 & \cellcolor[rgb]{0.437,0.686,0.216}82.84 & \cellcolor[rgb]{0.467,0.711,0.235}81.61 \\
 & \multirow{2}{*}{9.0} & PT & \cellcolor[rgb]{0.452,0.699,0.225}81.52 & \cellcolor[rgb]{0.586,0.794,0.361}76.03 & \cellcolor[rgb]{0.421,0.673,0.206}83.55 & \cellcolor[rgb]{0.421,0.673,0.206}83.49 \\
 &  & base & \cellcolor[rgb]{0.507,0.743,0.266}79.44 & \cellcolor[rgb]{0.533,0.760,0.297}78.24 & \cellcolor[rgb]{0.398,0.653,0.191}84.87 & \cellcolor[rgb]{0.406,0.660,0.196}84.50 \\
 & \multirow{2}{*}{35.0} & PT & \cellcolor[rgb]{0.437,0.686,0.216}82.44 & \cellcolor[rgb]{0.483,0.724,0.245}80.48 & \cellcolor[rgb]{0.390,0.647,0.186}84.89 & \cellcolor[rgb]{0.398,0.653,0.191}84.59 \\
 &  & base & \cellcolor[rgb]{0.429,0.679,0.211}83.45 & \cellcolor[rgb]{0.460,0.705,0.230}81.91 & \cellcolor[rgb]{0.390,0.647,0.186}85.47 & \cellcolor[rgb]{0.383,0.640,0.181}85.56 \\
\arrayrulecolor{lightgray}\cmidrule(lr){1-7}\arrayrulecolor{black}
\multirow{2}{*}{TinyAya} & \multirow{2}{*}{3.35} & PT & \cellcolor[rgb]{0.630,0.823,0.414}74.61 & \cellcolor[rgb]{0.936,0.965,0.894}54.77 & \cellcolor[rgb]{0.594,0.800,0.372}76.65 & \cellcolor[rgb]{0.612,0.811,0.393}75.52 \\
 &  & base & \cellcolor[rgb]{0.568,0.783,0.340}76.68 & \cellcolor[rgb]{0.936,0.965,0.894}55.01 & \cellcolor[rgb]{0.542,0.766,0.308}78.06 & \cellcolor[rgb]{0.551,0.771,0.319}77.51 \\
\bottomrule
\end{tabular}
\caption{Model accuracy by method, size, and base/PT variant.}
\label{tab:model_results_all_models}
\end{table}

\clearpage

\section{TinyAya Performance}\label{sec:tiny_aya}
\begin{figure}[H]
    \centering
    \includegraphics[]{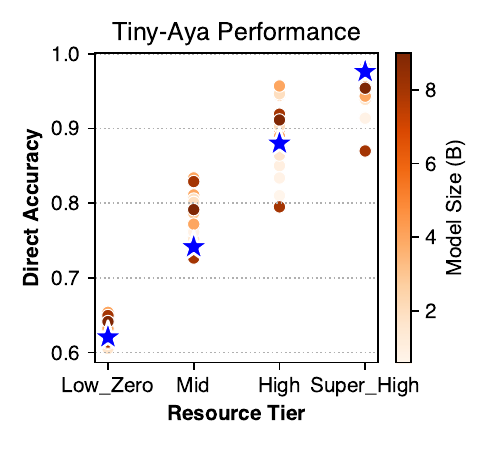}
    \caption{Performance of TinyAya (blue star) against all other models up to 9B parameters. TinyAya demonstrates a strong skew towards the Super High resource languages, and underperforms on low- and mid-resource languages.}
    \label{fig:tiny_aya}
\end{figure}
\clearpage

\section{Cross-Family Results}
\label{sec:appendix_families}

Tables~\ref{tab:appendix_base} and~\ref{tab:appendix_pt} report per-language Mean Direct accuracy for all model variants.

\makeatletter\@twocolumnfalse\makeatother
\setlength{\hsize}{\textwidth}%
\setlength{\linewidth}{\textwidth}%
\fontsize{5}{6}\selectfont
\setlength{\tabcolsep}{0.5pt}
\begin{longtable}{@{}llrrrrrrrrrrrrrrrrrrrrrrr@{}}
\caption{Mean Direct accuracy (\%) — Base models.}\\
\label{tab:appendix_base}\\
 & \multicolumn{4}{c}{Gemma~3} & \multicolumn{4}{c}{Gemma~4} & \multicolumn{3}{c}{Llama~3} & \multicolumn{6}{c}{Qwen~3} & \multicolumn{5}{c}{Qwen~3.5} & \multicolumn{1}{c}{TinyAya} \\
ISO & N & 1B & 4B & 12B & 27B & E2B & E4B & 26B-A4B & 31B & 1B & 3B & 8B & 0.6B & 1.7B & 4B & 8B & 14B & 30B-A3B & 0.8B & 2B & 4B & 9B & 35B-A3B & 3.35B \\
\midrule \endfirsthead
\multicolumn{25}{c}{{continued}} \\
ISO & N & 1B & 4B & 12B & 27B & E2B & E4B & 26B-A4B & 31B & 1B & 3B & 8B & 0.6B & 1.7B & 4B & 8B & 14B & 30B-A3B & 0.8B & 2B & 4B & 9B & 35B-A3B & 3.35B \\
\midrule \endhead
\midrule \multicolumn{25}{r}{{continued}} \\ \endfoot
\bottomrule \endlastfoot
abk & 40 & \cellcolor[rgb]{0.746,0.814,0.463}57.5 & \cellcolor[rgb]{0.736,0.812,0.450}60.0 & \cellcolor[rgb]{0.775,0.820,0.500}50.0 & \cellcolor[rgb]{0.658,0.796,0.350}80.0 & \cellcolor[rgb]{0.677,0.800,0.375}75.0 & \cellcolor[rgb]{0.658,0.796,0.350}80.0 & \cellcolor[rgb]{0.746,0.814,0.463}57.5 & \cellcolor[rgb]{0.775,0.820,0.500}50.0 & \cellcolor[rgb]{0.677,0.800,0.375}75.0 & \cellcolor[rgb]{0.697,0.804,0.400}70.0 & \cellcolor[rgb]{0.746,0.814,0.463}57.5 & \cellcolor[rgb]{0.765,0.818,0.487}52.5 & \cellcolor[rgb]{0.726,0.810,0.438}62.5 & \cellcolor[rgb]{0.785,0.822,0.512}47.5 & \cellcolor[rgb]{0.707,0.806,0.412}67.5 & \cellcolor[rgb]{0.755,0.816,0.475}55.0 & \cellcolor[rgb]{0.658,0.796,0.350}80.0 & \cellcolor[rgb]{0.677,0.800,0.375}75.0 & \cellcolor[rgb]{0.697,0.804,0.400}70.0 & \cellcolor[rgb]{0.746,0.814,0.463}57.5 & \cellcolor[rgb]{0.658,0.796,0.350}80.0 & \cellcolor[rgb]{0.804,0.826,0.537}42.5 & \cellcolor[rgb]{0.687,0.802,0.388}72.5 \\
aln & 677 & \cellcolor[rgb]{0.676,0.800,0.373}75.3 & \cellcolor[rgb]{0.675,0.799,0.372}75.6 & \cellcolor[rgb]{0.655,0.795,0.346}80.8 & \cellcolor[rgb]{0.651,0.795,0.341}81.8 & \cellcolor[rgb]{0.672,0.799,0.367}76.5 & \cellcolor[rgb]{0.658,0.796,0.350}79.9 & \cellcolor[rgb]{0.640,0.792,0.328}84.5 & \cellcolor[rgb]{0.632,0.791,0.317}86.6 & \cellcolor[rgb]{0.693,0.803,0.396}70.9 & \cellcolor[rgb]{0.698,0.804,0.401}69.9 & \cellcolor[rgb]{0.679,0.800,0.377}74.6 & \cellcolor[rgb]{0.696,0.804,0.398}70.3 & \cellcolor[rgb]{0.692,0.803,0.393}71.3 & \cellcolor[rgb]{0.680,0.800,0.378}74.5 & \cellcolor[rgb]{0.699,0.804,0.403}69.4 & \cellcolor[rgb]{0.697,0.804,0.400}70.0 & \cellcolor[rgb]{0.683,0.801,0.381}73.7 & \cellcolor[rgb]{0.703,0.805,0.408}68.4 & \cellcolor[rgb]{0.689,0.802,0.390}72.1 & \cellcolor[rgb]{0.691,0.803,0.392}71.6 & \cellcolor[rgb]{0.677,0.800,0.374}75.2 & \cellcolor[rgb]{0.675,0.799,0.372}75.6 & \cellcolor[rgb]{0.681,0.801,0.379}74.2 \\
\end{longtable}

\clearpage
\makeatletter\@twocolumnfalse\makeatother
\setlength{\tabcolsep}{0.5pt}
\begin{longtable}{@{}llrrrrrrrrrrrrrrrrrrrrrrr@{}}
\caption{Mean Direct accuracy (\%) — PT models.}\\
\label{tab:appendix_pt}\\
 & \multicolumn{4}{c}{Gemma~3} & \multicolumn{4}{c}{Gemma~4} & \multicolumn{3}{c}{Llama~3} & \multicolumn{6}{c}{Qwen~3} & \multicolumn{5}{c}{Qwen~3.5} & \multicolumn{1}{c}{TinyAya} \\
ISO & N & 1B-it & 4B-it & 12B-it & 27B-it & E2B-it & E4B-it & 26B-A4B-it & 31B-it & 1B-it & 3B-it & 8B-it & 0.6B-it & 1.7B-it & 4B-it & 8B-it & 14B-it & 30B-A3B-it & 0.8B-it & 2B-it & 4B-it & 9B-it & 35B-A3B-it & 3.35B-it \\
\midrule \endfirsthead
\multicolumn{25}{c}{{continued}} \\
ISO & N & 1B-it & 4B-it & 12B-it & 27B-it & E2B-it & E4B-it & 26B-A4B-it & 31B-it & 1B-it & 3B-it & 8B-it & 0.6B-it & 1.7B-it & 4B-it & 8B-it & 14B-it & 30B-A3B-it & 0.8B-it & 2B-it & 4B-it & 9B-it & 35B-A3B-it & 3.35B-it \\
\midrule \endhead
\midrule \multicolumn{25}{r}{{continued}} \\ \endfoot
\bottomrule \endlastfoot
abk & 40 & \cellcolor[rgb]{0.677,0.800,0.375}75.0 & \cellcolor[rgb]{0.853,0.836,0.600}30.0 & \cellcolor[rgb]{0.785,0.822,0.512}47.5 & \cellcolor[rgb]{0.687,0.802,0.388}72.5 & \cellcolor[rgb]{0.726,0.810,0.438}62.5 & \cellcolor[rgb]{0.716,0.808,0.425}65.0 & \cellcolor[rgb]{0.687,0.802,0.388}72.5 & \cellcolor[rgb]{0.736,0.812,0.450}60.0 & \cellcolor[rgb]{0.687,0.802,0.388}72.5 & \cellcolor[rgb]{0.677,0.800,0.375}75.0 & \cellcolor[rgb]{0.687,0.802,0.388}72.5 & \cellcolor[rgb]{0.804,0.826,0.537}42.5 & \cellcolor[rgb]{0.794,0.824,0.525}45.0 & \cellcolor[rgb]{0.775,0.820,0.500}50.0 & \cellcolor[rgb]{0.687,0.802,0.388}72.5 & \cellcolor[rgb]{0.775,0.820,0.500}50.0 & \cellcolor[rgb]{0.697,0.804,0.400}70.0 & \cellcolor[rgb]{0.677,0.800,0.375}75.0 & \cellcolor[rgb]{0.726,0.810,0.438}62.5 & \cellcolor[rgb]{0.726,0.810,0.438}62.5 & \cellcolor[rgb]{0.707,0.806,0.412}67.5 & \cellcolor[rgb]{0.765,0.818,0.487}52.5 & \cellcolor[rgb]{0.736,0.812,0.450}60.0 \\
aln & 677 & \cellcolor[rgb]{0.727,0.810,0.438}62.3 & \cellcolor[rgb]{0.726,0.810,0.438}62.5 & \cellcolor[rgb]{0.676,0.800,0.373}75.3 & \cellcolor[rgb]{0.679,0.800,0.376}74.7 & \cellcolor[rgb]{0.717,0.808,0.425}65.0 & \cellcolor[rgb]{0.705,0.806,0.410}68.0 & \cellcolor[rgb]{0.714,0.808,0.422}65.6 & \cellcolor[rgb]{0.707,0.806,0.413}67.4 & \cellcolor[rgb]{0.704,0.806,0.410}68.1 & \cellcolor[rgb]{0.711,0.807,0.418}66.3 & \cellcolor[rgb]{0.709,0.806,0.415}66.9 & \cellcolor[rgb]{0.705,0.806,0.410}68.0 & \cellcolor[rgb]{0.703,0.805,0.407}68.5 & \cellcolor[rgb]{0.699,0.804,0.403}69.4 & \cellcolor[rgb]{0.696,0.804,0.398}70.3 & \cellcolor[rgb]{0.709,0.806,0.415}66.9 & \cellcolor[rgb]{0.704,0.805,0.409}68.2 & \cellcolor[rgb]{0.700,0.805,0.404}69.3 & \cellcolor[rgb]{0.685,0.802,0.385}73.0 & \cellcolor[rgb]{0.685,0.802,0.384}73.1 & \cellcolor[rgb]{0.681,0.801,0.379}74.2 & \cellcolor[rgb]{0.669,0.798,0.364}77.2 & \cellcolor[rgb]{0.692,0.803,0.393}71.3 \\
\end{longtable}
\normalsize

\subsection{CC-100 Frequency and Base Accuracy}

Table~\ref{tab:appendix_cc_frequency} reports within-family OLS
regression results predicting base accuracy from log-transformed CC-100
corpus size and model scale. The cross-family mixed-effects model
(with model as random intercept) is reported in the last two rows of the
table.

\begin{table}[H]
\centering
\caption{Within-family OLS: $\mathrm{Accuracy} \sim \log_{10}(\mathrm{CC\ count}) + \mathrm{Size\_Order}$.
$\mathrm{Size\_Order}$ is the rank of the model size within its family, starting at~0.
All models are base, Direct method.}
\label{tab:appendix_cc_frequency}
\setlength{\tabcolsep}{6pt}
\begin{tabular}{@{}lrrrr@{}}
\toprule
Family & \multicolumn{1}{c}{$N$} & \multicolumn{1}{c}{$\beta_{\mathrm{CC}}$} & \multicolumn{1}{c}{$p$} & \multicolumn{1}{c}{$R^2$} \\
\midrule
Gemma~3      & 356 & 0.0610 & $<0.001$ & 0.651 \\
Gemma~4      & 356 & 0.0613 & $<0.001$ & 0.630 \\
Llama~3.1    &  89 & 0.0578 & $<0.001$ & 0.629 \\
Llama~3.2    & 178 & 0.0535 & $<0.001$ & 0.569 \\
Qwen~3       & 534 & 0.0552 & $<0.001$ & 0.535 \\
Qwen~3.5     & 445 & 0.0555 & $<0.001$ & 0.591 \\
TinyAya      &  89 & 0.0619 & $<0.001$ & 0.538 \\
\midrule
\multicolumn{5}{l}{\textit{Cross-family MixedLM (model as random intercept):}} \\
\multicolumn{5}{l}{$\beta_{\mathrm{CC}} = 0.059$, $p < 0.001$; ICC $= 0.078$} \\
\bottomrule
\end{tabular}
\end{table}

\subsection{Log-Probability Differences}

\begin{figure}[H]
    \centering
    \resizebox{\columnwidth}{!}{\begin{tikzpicture}
\begin{axis}[
    width=12cm, height=8cm,
    xlabel={Direct accuracy (\%)},
    ylabel={Mean $\log P(S_{cor}) - \log P(S_{inc})$},
    xmin=8, xmax=102,
    ymin=-8, ymax=18,
    legend pos=south east,
    legend style={font=\small},
    axis lines=left,
    clip=false,
]
\addplot[black!30, dashed, thin, forget plot] coordinates {(8,0) (102,0)};
\addplot[blue!70!black, only marks, mark=*, mark size=1.7pt, fill opacity=0.7, draw opacity=0.9]
    table[col sep=space, x=acc, y=delta] {figures/data/delta_acc_gemma3_Super_High.dat};
\addplot[cyan!60!blue, only marks, mark=*, mark size=1.7pt, fill opacity=0.7, draw opacity=0.9]
    table[col sep=space, x=acc, y=delta] {figures/data/delta_acc_gemma3_High.dat};
\addplot[orange!80!red, only marks, mark=*, mark size=1.7pt, fill opacity=0.7, draw opacity=0.9]
    table[col sep=space, x=acc, y=delta] {figures/data/delta_acc_gemma3_Mid.dat};
\addplot[red!80!black, only marks, mark=*, mark size=1.7pt, fill opacity=0.7, draw opacity=0.9]
    table[col sep=space, x=acc, y=delta] {figures/data/delta_acc_gemma3_Low_Zero.dat};
\legend{Super High, High, Mid, Low/Zero}
\end{axis}
\end{tikzpicture}}
    \caption{Direct accuracy against the mean log-probability difference between the grammatical and ungrammatical sentence, one point per language, for Gemma~3 (27B, base). Colour indicates resource tier. The dashed line marks equal probability.}
    \label{fig:delta_acc_gemma3}
\end{figure}
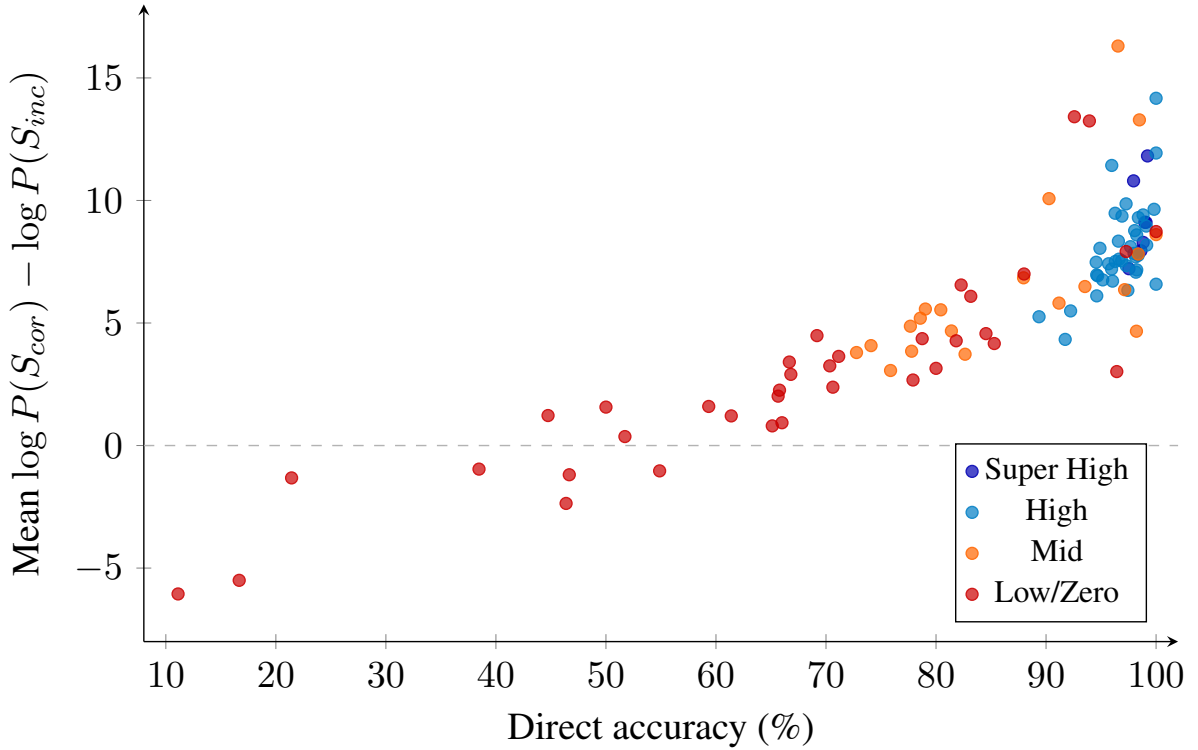

Binary accuracy records only which of the two sentences receives the higher log probability, not by how much. Figure~\ref{fig:delta_acc_gemma3} plots, for each of the 101 languages, the Direct accuracy of Gemma~3 (27B, base) against the mean of $\log P(S_{cor}) - \log P(S_{inc})$ over that language's items. Averaged within tiers, this difference is 9.18, 8.02, 6.53 and 2.85 natural-log units from Super High to Low/Zero. Five Low/Zero languages have a negative mean, meaning the model assigns higher probability to the ungrammatical sentence on average; the smallest are Tupinamb\'a ($-6.06$, 9 items) and Warlpiri ($-5.50$, 12 items).

Table~\ref{tab:appendix_delta_tier} reports the same quantity at the item level. The upper end of the distribution is almost identical across tiers (p90 of 15.0, 14.0, 14.0 and 14.0), so the items a model separates confidently are separated just as widely in Low/Zero as in Super High. The tiers differ at the lower end: p10 falls from 3.5 to $-5.0$, and the share of items separated by at most one natural-log unit rises from 2.6\% to 14.2\%.

\begin{table}[H]
\centering
\small
\caption{Distribution of the per-item log-probability difference $\log P(S_{cor}) - \log P(S_{inc})$ by resource tier, for Gemma~3 (27B, base) under Direct evaluation. Percentiles are taken over items. The last column is the share of items on which the two sentences are separated by at most one natural-log unit in either direction.}
\label{tab:appendix_delta_tier}
\setlength{\tabcolsep}{4pt}
\begin{tabular}{@{}lrrrrrrr@{}}
\toprule
Tier & Items & p10 & p25 & Median & p75 & p90 & Near-tie \\
\midrule
Super High & 18,260 & 3.5 & 5.5 & 8.0 & 11.0 & 15.0 & 2.6\% \\
High & 65,010 & 2.0 & 4.5 & 7.0 & 10.0 & 14.0 & 5.1\% \\
Mid & 23,936 & $-2.0$ & 1.0 & 5.0 & 8.5 & 14.0 & 12.9\% \\
Low/Zero & 14,099 & $-5.0$ & 0.0 & 4.0 & 8.0 & 14.0 & 14.2\% \\
\bottomrule
\end{tabular}
\end{table}

\subsection{Raw Alignment Tax}

Table~\ref{tab:appendix_raw_tax} reports the alignment tax in raw percentage points, without the normalization by above-chance base accuracy used in \S\ref{sec:alignment_tax}. In raw terms the tax is not systematically largest in the Low/Zero tier: for Gemma~4 it peaks in the Mid tier, and for TinyAya it is smallest in Low/Zero. Base accuracy in that tier is already close to the 50\% chance level, so there are few percentage points left to lose; dividing by $\mathrm{Acc}_{\mathrm{base}} - 0.5$ is what makes the tax comparable across tiers.

\begin{table}[H]
\centering
\caption{Raw linguistic alignment tax, $\mathrm{Acc}_{\mathrm{base}} - \mathrm{Acc}_{\mathrm{PT}}$ in percentage points, by resource tier for the largest model in each family (Direct method). Unlike the normalized tax, no languages are excluded; values are averaged equally across the languages in each tier (7, 38, 20 and 36 languages) and across all 101 languages in the last column.}
\label{tab:appendix_raw_tax}
\setlength{\tabcolsep}{5pt}
\begin{tabular}{@{}lrrrrr@{}}
\toprule
Family & Super High & High & Mid & Low/Zero & All \\
\midrule
Gemma~3 (27B) & 3.45 & 4.28 & 5.40 & 5.02 & 4.71 \\
Gemma~4 (31B) & 7.07 & 10.81 & 11.27 & 7.53 & 9.47 \\
Llama~3 (8B) & 1.80 & 2.03 & 1.25 & 1.54 & 1.69 \\
Qwen~3 (30B-A3B) & 2.38 & 2.30 & 3.05 & 1.68 & 2.23 \\
Qwen~3.5 (35B-A3B) & 0.94 & 1.20 & 1.38 & 0.63 & 1.02 \\
TinyAya (3.35B) & 4.40 & 2.90 & 2.95 & 0.26 & 2.07 \\
\bottomrule
\end{tabular}
\end{table}

\subsection{Small Test Sets in the Low/Zero Tier}
\label{sec:appendix_smalln}

The Low/Zero tier contains the smallest test sets in MultiBLiMP, and a mean
over languages weights a 26-item language as heavily as a 4{,}166-item one.
Table~\ref{tab:appendix_smalln} repeats the Gemma~3 tax of
\S\ref{sec:alignment_tax} while dropping the smallest test sets one at a time,
down to half of the tier. Each row uses a subset of the languages of the row
above it, on the same models and the same items, so the rows are directly
comparable.

\begin{table}[H]
\centering
\small
\caption{Low/Zero normalized alignment tax for Gemma~3 (Direct method) after
repeatedly dropping the language with the smallest test set. Row $k$ uses the
$24-k$ Low/Zero languages with the largest test sets, so every row is a subset
of the row above it; the models and the items are otherwise unchanged.
``Smallest $N$'' is the size of the smallest test set still included. Cells are
the mean over the retained languages with a 95\% bootstrap confidence interval
(10{,}000 resamples). The 12 Low/Zero languages with $\mathrm{Acc}_{\mathrm{base}}
\leq 0.5$ are excluded from all rows, as in \S\ref{sec:alignment_tax}.}
\label{tab:appendix_smalln}
\setlength{\tabcolsep}{6pt}
\begin{tabular}{@{}rrrcccc@{}}
\toprule
Dropped & $n$ & Smallest $N$ & 1B & 4B & 12B & 27B \\
\midrule
 0 & 24 &  26 & 35.9 [24.2, 47.4] & 33.0 [21.2, 43.7] & 12.0 [$-$10.7, 28.1] & 20.6 [12.8, 28.1] \\
 1 & 23 &  28 & 36.7 [24.8, 48.5] & 32.7 [20.6, 44.2] & 10.5 [$-$12.4, 26.9] & 20.1 [12.2, 27.9] \\
 2 & 22 &  34 & 34.6 [23.1, 46.4] & 33.2 [20.5, 44.7] &  9.5 [$-$14.2, 26.9] & 18.6 [10.9, 26.0] \\
 3 & 21 &  50 & 32.0 [20.9, 43.2] & 37.5 [28.8, 47.1] &  8.4 [$-$17.6, 26.7] & 17.5 [\phantom{0}9.6, 25.2] \\
 4 & 20 &  86 & 34.2 [23.2, 45.0] & 39.6 [31.7, 48.7] &  8.4 [$-$18.3, 27.3] & 17.3 [\phantom{0}9.2, 25.4] \\
 5 & 19 &  99 & 35.2 [23.3, 46.4] & 40.1 [31.9, 50.0] &  7.9 [$-$20.7, 27.7] & 18.7 [10.3, 26.8] \\
 6 & 18 & 131 & 35.1 [23.0, 47.0] & 41.4 [33.3, 51.7] &  7.9 [$-$21.4, 28.8] & 19.7 [11.1, 27.9] \\
 7 & 17 & 187 & 36.1 [23.4, 48.3] & 42.2 [33.3, 52.8] & 11.9 [$-$18.5, 31.5] & 20.6 [11.6, 28.9] \\
 8 & 16 & 190 & 36.9 [23.5, 49.8] & 42.7 [33.3, 54.1] & 10.2 [$-$21.3, 30.4] & 18.5 [10.0, 26.2] \\
 9 & 15 & 207 & 34.7 [20.9, 47.4] & 40.9 [31.8, 52.5] &  9.8 [$-$24.4, 31.5] & 17.2 [\phantom{0}8.6, 24.9] \\
10 & 14 & 241 & 34.8 [20.3, 48.9] & 41.8 [32.2, 53.9] & 10.5 [$-$26.4, 33.5] & 17.4 [\phantom{0}8.2, 25.7] \\
11 & 13 & 254 & 39.3 [27.1, 51.3] & 43.5 [33.8, 55.8] &  8.7 [$-$29.4, 33.5] & 18.7 [\phantom{0}8.9, 27.1] \\
12 & 12 & 260 & 37.8 [25.6, 51.1] & 43.1 [32.7, 56.2] &  5.7 [$-$35.6, 31.9] & 18.2 [\phantom{0}7.7, 27.2] \\
\bottomrule
\end{tabular}
\end{table}

At 27B the tax stays between 17.2\% and 20.6\% across the entire sweep, and
Low/Zero remains the highest-tax tier at every cutoff (the other tiers are
untouched at 7.1\%, 9.3\% and 16.1\%), although its confidence interval
overlaps the Mid tier's throughout. The 12B estimate is the unstable one: its
mean moves between 5.7\% and 12.0\% and its interval always includes zero,
which is why we treat the 12B$\rightarrow$27B rebound in
\S\ref{sec:alignment_tax} as within uncertainty. That rebound stays
non-significant under every cutoff (paired difference $+6.9$ to $+12.4$pp,
$p$ from $0.26$ to $0.68$). The 4B estimate rises from 33.0\% to 43.1\% as
small test sets are dropped, so the small languages were pulling that
particular figure down rather than up.

Dropping the smallest test sets also raises Low/Zero Direct accuracy in
Table~\ref{tab:lang_tier_accuracy}: for Gemma~3 (27B) the base/post-trained
pair moves from 66.7/61.7 with all 36 Low/Zero languages to 72.0/67.9 with the
12 largest. The base-above-post-trained ordering holds for all six families at
every cutoff.

\subsection{Gap by Resource Tier and Family}

Table~\ref{tab:appendix_tier_gap} reports the mean inarticulate gap
for the largest model in each family, broken down by resource tier.

\begin{table}[H]
\centering
\caption{Mean inarticulate gap by resource tier and family (largest
model only).}
\label{tab:appendix_tier_gap}
\setlength{\tabcolsep}{6pt}
\begin{tabular}{@{}lrrrr@{}}
\toprule
Family & Super High & High & Mid & Low/Zero \\
\midrule
Gemma~3 (27B)  & 0.038 & 0.044 & 0.094 & 0.047 \\
Gemma~4 (31B)  & $-0.066$ & $-0.091$ & $-0.044$ & $-0.042$ \\
Llama~3 (8B) & 0.233 & 0.272 & 0.292 & 0.120 \\
Qwen~3 (30B-A3B) & 0.074 & 0.088 & 0.133 & 0.085 \\
Qwen~3.5 (35B-A3B) & 0.003 & 0.004 & 0.020 & 0.039 \\
TinyAya (3.35B) & 0.277 & 0.244 & 0.209 & 0.129 \\
\bottomrule
\end{tabular}
\end{table}

\begin{table}[htbp]
\centering
\setlength{\tabcolsep}{6pt}
\scriptsize
\begin{tabular}{@{}lll>{\centering\arraybackslash}p{0.8cm}>{\centering\arraybackslash}p{0.8cm}>{\centering\arraybackslash}p{0.8cm}>{\centering\arraybackslash}p{0.8cm}}
& & & \multicolumn{4}{c}{\textit{Resource Tier}} \\
\arrayrulecolor{lightgray}\cmidrule(lr){4-7}\arrayrulecolor{black}
 & Size & Variant & \textbf{Low Zero} & \textbf{Mid} & \textbf{High} & \textbf{Super High} \\
\midrule
\multirow{8}{*}{Gemma~3} & \multirow{2}{*}{1.0} & PT & \cellcolor[rgb]{0.667,0.798,0.362}3.50 & \cellcolor[rgb]{0.749,0.815,0.467}15.32 & \cellcolor[rgb]{0.780,0.821,0.507}19.83 & \cellcolor[rgb]{0.801,0.825,0.533}22.85 \\
 &  & base & \cellcolor[rgb]{0.727,0.810,0.438}12.13 & \cellcolor[rgb]{0.809,0.827,0.543}23.98 & \cellcolor[rgb]{0.927,0.851,0.694}40.97 & \cellcolor[rgb]{0.970,0.860,0.750}47.25 \\
 & \multirow{2}{*}{4.0} & PT & \cellcolor[rgb]{0.686,0.802,0.386}6.20 & \cellcolor[rgb]{0.727,0.810,0.438}12.08 & \cellcolor[rgb]{0.712,0.807,0.419}9.91 & \cellcolor[rgb]{0.698,0.804,0.401}7.93 \\
 &  & base & \cellcolor[rgb]{0.723,0.809,0.433}11.52 & \cellcolor[rgb]{0.842,0.834,0.586}28.72 & \cellcolor[rgb]{0.821,0.829,0.559}25.67 & \cellcolor[rgb]{0.833,0.832,0.574}27.41 \\
 & \multirow{2}{*}{12.0} & PT & \cellcolor[rgb]{0.674,0.799,0.371}4.49 & \cellcolor[rgb]{0.716,0.808,0.424}10.54 & \cellcolor[rgb]{0.673,0.799,0.369}4.32 & \cellcolor[rgb]{0.686,0.802,0.386}6.22 \\
 &  & base & \cellcolor[rgb]{0.738,0.812,0.452}13.70 & \cellcolor[rgb]{0.809,0.827,0.544}24.00 & \cellcolor[rgb]{0.775,0.820,0.500}19.13 & \cellcolor[rgb]{0.725,0.810,0.435}11.79 \\
 & \multirow{2}{*}{27.0} & PT & \cellcolor[rgb]{0.675,0.800,0.372}4.66 & \cellcolor[rgb]{0.708,0.806,0.414}9.36 & \cellcolor[rgb]{0.674,0.799,0.370}4.43 & \cellcolor[rgb]{0.669,0.798,0.365}3.81 \\
 &  & base & \cellcolor[rgb]{0.725,0.810,0.436}11.86 & \cellcolor[rgb]{0.755,0.816,0.475}16.22 & \cellcolor[rgb]{0.700,0.805,0.403}8.18 & \cellcolor[rgb]{0.691,0.803,0.393}6.97 \\
\arrayrulecolor{lightgray}\cmidrule(lr){1-7}\arrayrulecolor{black}
\multirow{8}{*}{Gemma~4} & \multirow{2}{*}{2.0} & PT & \cellcolor[rgb]{0.690,0.803,0.391}6.78 & \cellcolor[rgb]{0.753,0.816,0.472}15.91 & \cellcolor[rgb]{0.700,0.805,0.404}8.28 & \cellcolor[rgb]{0.697,0.804,0.400}7.76 \\
 &  & base & \cellcolor[rgb]{0.737,0.812,0.452}13.66 & \cellcolor[rgb]{0.851,0.836,0.598}30.11 & \cellcolor[rgb]{0.952,0.856,0.727}44.67 & \cellcolor[rgb]{0.950,0.856,0.724}44.34 \\
 & \multirow{2}{*}{4.0} & PT & \cellcolor[rgb]{0.696,0.804,0.399}7.68 & \cellcolor[rgb]{0.758,0.816,0.478}16.57 & \cellcolor[rgb]{0.734,0.812,0.447}13.10 & \cellcolor[rgb]{0.707,0.806,0.413}9.23 \\
 &  & base & \cellcolor[rgb]{0.736,0.812,0.450}13.43 & \cellcolor[rgb]{0.772,0.819,0.496}18.66 & \cellcolor[rgb]{0.795,0.824,0.526}21.99 & \cellcolor[rgb]{0.752,0.815,0.471}15.80 \\
 & \multirow{2}{*}{26.0} & PT & \cellcolor[rgb]{0.656,0.795,0.347}1.81 & \cellcolor[rgb]{0.633,0.791,0.318}-1.41 & \cellcolor[rgb]{0.585,0.781,0.256}-8.38 & \cellcolor[rgb]{0.603,0.785,0.279}-5.79 \\
 &  & base & \cellcolor[rgb]{0.701,0.805,0.405}8.38 & \cellcolor[rgb]{0.718,0.808,0.427}10.84 & \cellcolor[rgb]{0.688,0.802,0.388}6.51 & \cellcolor[rgb]{0.675,0.800,0.372}4.65 \\
 & \multirow{2}{*}{31.0} & PT & \cellcolor[rgb]{0.614,0.787,0.293}\textbf{-4.20} & \cellcolor[rgb]{0.613,0.787,0.292}\textbf{-4.38} & \cellcolor[rgb]{0.580,0.780,0.250}\textbf{-9.10} & \cellcolor[rgb]{0.597,0.784,0.272}\textbf{-6.61} \\
 &  & base & \cellcolor[rgb]{0.639,0.792,0.325}-0.59 & \cellcolor[rgb]{0.669,0.798,0.363}3.69 & \cellcolor[rgb]{0.655,0.795,0.346}1.70 & \cellcolor[rgb]{0.645,0.793,0.334}0.33 \\
\arrayrulecolor{lightgray}\cmidrule(lr){1-7}\arrayrulecolor{black}
\multirow{6}{*}{Llama~3} & \multirow{2}{*}{1.0} & PT & \cellcolor[rgb]{0.708,0.806,0.414}9.38 & \cellcolor[rgb]{0.805,0.826,0.538}23.35 & \cellcolor[rgb]{0.841,0.834,0.585}28.63 & \cellcolor[rgb]{0.902,0.846,0.663}37.50 \\
 &  & base & \cellcolor[rgb]{0.725,0.810,0.436}11.88 & \cellcolor[rgb]{0.830,0.831,0.570}27.01 & \cellcolor[rgb]{0.891,0.844,0.649}35.89 & \cellcolor[rgb]{0.956,0.857,0.732}45.22 \\
 & \multirow{2}{*}{3.0} & PT & \cellcolor[rgb]{0.702,0.805,0.406}8.47 & \cellcolor[rgb]{0.812,0.828,0.547}24.40 & \cellcolor[rgb]{0.839,0.833,0.582}28.28 & \cellcolor[rgb]{0.851,0.836,0.598}30.07 \\
 &  & base & \cellcolor[rgb]{0.745,0.814,0.461}14.73 & \cellcolor[rgb]{0.837,0.833,0.579}28.03 & \cellcolor[rgb]{0.901,0.846,0.662}37.30 & \cellcolor[rgb]{0.930,0.852,0.698}41.42 \\
 & \multirow{2}{*}{8.0} & PT & \cellcolor[rgb]{0.726,0.810,0.437}12.00 & \cellcolor[rgb]{0.845,0.834,0.590}29.17 & \cellcolor[rgb]{0.831,0.831,0.572}27.15 & \cellcolor[rgb]{0.804,0.826,0.537}23.28 \\
 &  & base & \cellcolor[rgb]{0.740,0.813,0.455}14.00 & \cellcolor[rgb]{0.858,0.837,0.607}31.09 & \cellcolor[rgb]{0.878,0.841,0.632}33.98 & \cellcolor[rgb]{0.887,0.843,0.643}35.24 \\
\arrayrulecolor{lightgray}\cmidrule(lr){1-7}\arrayrulecolor{black}
\multirow{12}{*}{Qwen~3} & \multirow{2}{*}{0.6} & PT & \cellcolor[rgb]{0.717,0.808,0.426}10.75 & \cellcolor[rgb]{0.792,0.823,0.522}21.53 & \cellcolor[rgb]{0.822,0.830,0.560}25.82 & \cellcolor[rgb]{0.866,0.839,0.617}32.29 \\
 &  & base & \cellcolor[rgb]{0.725,0.810,0.436}11.84 & \cellcolor[rgb]{0.795,0.824,0.525}21.92 & \cellcolor[rgb]{0.846,0.835,0.591}29.38 & \cellcolor[rgb]{0.907,0.847,0.669}38.08 \\
 & \multirow{2}{*}{1.7} & PT & \cellcolor[rgb]{0.698,0.804,0.401}7.88 & \cellcolor[rgb]{0.777,0.820,0.503}19.41 & \cellcolor[rgb]{0.800,0.825,0.532}22.69 & \cellcolor[rgb]{0.768,0.819,0.491}18.11 \\
 &  & base & \cellcolor[rgb]{0.716,0.808,0.425}10.59 & \cellcolor[rgb]{0.805,0.826,0.539}23.44 & \cellcolor[rgb]{0.809,0.827,0.543}23.98 & \cellcolor[rgb]{0.797,0.824,0.528}22.24 \\
 & \multirow{2}{*}{4.0} & PT & \cellcolor[rgb]{0.707,0.806,0.413}9.28 & \cellcolor[rgb]{0.814,0.828,0.551}24.77 & \cellcolor[rgb]{0.841,0.834,0.585}28.64 & \cellcolor[rgb]{0.870,0.840,0.622}32.81 \\
 &  & base & \cellcolor[rgb]{0.675,0.799,0.372}4.63 & \cellcolor[rgb]{0.776,0.820,0.502}19.29 & \cellcolor[rgb]{0.764,0.818,0.485}17.42 & \cellcolor[rgb]{0.724,0.810,0.435}11.77 \\
 & \multirow{2}{*}{8.0} & PT & \cellcolor[rgb]{0.731,0.811,0.444}12.72 & \cellcolor[rgb]{0.832,0.832,0.573}27.32 & \cellcolor[rgb]{0.911,0.848,0.675}38.78 & \cellcolor[rgb]{0.943,0.854,0.715}43.29 \\
 &  & base & \cellcolor[rgb]{0.674,0.799,0.371}4.54 & \cellcolor[rgb]{0.701,0.805,0.405}8.42 & \cellcolor[rgb]{0.624,0.789,0.306}-2.71 & \cellcolor[rgb]{0.621,0.789,0.302}-3.15 \\
 & \multirow{2}{*}{14.0} & PT & \cellcolor[rgb]{0.670,0.799,0.366}3.97 & \cellcolor[rgb]{0.723,0.809,0.433}11.57 & \cellcolor[rgb]{0.687,0.802,0.388}6.40 & \cellcolor[rgb]{0.677,0.800,0.374}4.88 \\
 &  & base & \cellcolor[rgb]{0.680,0.800,0.378}5.32 & \cellcolor[rgb]{0.714,0.808,0.422}10.32 & \cellcolor[rgb]{0.680,0.800,0.378}5.31 & \cellcolor[rgb]{0.661,0.797,0.354}2.66 \\
 & \multirow{2}{*}{30.0} & PT & \cellcolor[rgb]{0.702,0.805,0.406}8.47 & \cellcolor[rgb]{0.735,0.812,0.448}13.25 & \cellcolor[rgb]{0.704,0.805,0.409}8.82 & \cellcolor[rgb]{0.694,0.803,0.396}7.40 \\
 &  & base & \cellcolor[rgb]{0.694,0.803,0.397}7.44 & \cellcolor[rgb]{0.732,0.811,0.445}12.89 & \cellcolor[rgb]{0.702,0.805,0.406}8.53 & \cellcolor[rgb]{0.698,0.804,0.402}7.98 \\
\arrayrulecolor{lightgray}\cmidrule(lr){1-7}\arrayrulecolor{black}
\multirow{10}{*}{Qwen~3.5} & \multirow{2}{*}{0.8} & PT & \cellcolor[rgb]{0.727,0.810,0.439}12.17 & \cellcolor[rgb]{0.796,0.824,0.527}22.13 & \cellcolor[rgb]{0.812,0.827,0.547}24.36 & \cellcolor[rgb]{0.851,0.836,0.597}30.00 \\
 &  & base & \cellcolor[rgb]{0.725,0.810,0.436}11.87 & \cellcolor[rgb]{0.791,0.823,0.520}21.33 & \cellcolor[rgb]{0.793,0.824,0.523}21.63 & \cellcolor[rgb]{0.804,0.826,0.537}23.27 \\
 & \multirow{2}{*}{2.0} & PT & \cellcolor[rgb]{0.722,0.809,0.433}11.48 & \cellcolor[rgb]{0.801,0.825,0.533}22.79 & \cellcolor[rgb]{0.763,0.818,0.485}17.37 & \cellcolor[rgb]{0.745,0.814,0.461}14.72 \\
 &  & base & \cellcolor[rgb]{0.717,0.808,0.425}10.67 & \cellcolor[rgb]{0.790,0.823,0.519}21.21 & \cellcolor[rgb]{0.774,0.820,0.498}18.88 & \cellcolor[rgb]{0.765,0.818,0.488}17.69 \\
 & \multirow{2}{*}{4.0} & PT & \cellcolor[rgb]{0.703,0.805,0.408}8.68 & \cellcolor[rgb]{0.738,0.812,0.452}13.69 & \cellcolor[rgb]{0.720,0.809,0.429}11.09 & \cellcolor[rgb]{0.718,0.808,0.428}10.91 \\
 &  & base & \cellcolor[rgb]{0.688,0.802,0.388}6.48 & \cellcolor[rgb]{0.700,0.805,0.404}8.24 & \cellcolor[rgb]{0.684,0.801,0.383}5.88 & \cellcolor[rgb]{0.677,0.800,0.374}4.88 \\
 & \multirow{2}{*}{9.0} & PT & \cellcolor[rgb]{0.683,0.801,0.382}5.75 & \cellcolor[rgb]{0.681,0.801,0.379}5.49 & \cellcolor[rgb]{0.680,0.801,0.378}5.37 & \cellcolor[rgb]{0.677,0.800,0.374}4.85 \\
 &  & base & \cellcolor[rgb]{0.677,0.800,0.374}4.88 & \cellcolor[rgb]{0.632,0.791,0.316}-1.61 & \cellcolor[rgb]{0.639,0.792,0.325}-0.63 & \cellcolor[rgb]{0.644,0.793,0.332}0.19 \\
 & \multirow{2}{*}{35.0} & PT & \cellcolor[rgb]{0.670,0.798,0.365}3.90 & \cellcolor[rgb]{0.657,0.796,0.349}2.02 & \cellcolor[rgb]{0.646,0.793,0.334}0.39 & \cellcolor[rgb]{0.645,0.793,0.333}0.30 \\
 &  & base & \cellcolor[rgb]{0.657,0.796,0.348}1.97 & \cellcolor[rgb]{0.663,0.797,0.356}2.83 & \cellcolor[rgb]{0.647,0.794,0.336}0.56 & \cellcolor[rgb]{0.651,0.794,0.340}1.10 \\
\arrayrulecolor{lightgray}\cmidrule(lr){1-7}\arrayrulecolor{black}
\multirow{2}{*}{TinyAya} & \multirow{2}{*}{3.35} & PT & \cellcolor[rgb]{0.732,0.811,0.445}12.91 & \cellcolor[rgb]{0.788,0.823,0.517}20.94 & \cellcolor[rgb]{0.812,0.828,0.547}24.37 & \cellcolor[rgb]{0.835,0.832,0.577}27.70 \\
 &  & base & \cellcolor[rgb]{0.712,0.807,0.419}9.97 & \cellcolor[rgb]{0.807,0.827,0.541}23.68 & \cellcolor[rgb]{0.848,0.835,0.593}29.59 & \cellcolor[rgb]{0.872,0.840,0.624}33.08 \\
\bottomrule
\end{tabular}
\caption{Mean inarticulate gap (Acc$_{\mathrm{Direct}} -$ Acc$_{\mathrm{Meta}}$, \%) by resource tier for all model variants. Bold=lowest gap per column.}
\label{tab:base_pt_gap}
\end{table}

\subsection{Substitute Native Cues and the Prompt Gain}
\label{sec:appendix_subcue}

For 14 of the 101 languages the native cue is not a translation of the English
template (Appendix~\ref{sec:appendix_native_prompts}); 13 of these fall in the
Low/Zero tier. This raises the question of whether the Prompt Gain reported for
that tier in \S\ref{sec:language_prompting} is an artefact of those cues.

Across the 23 post-trained models the Low/Zero tier splits into 23 languages
with a fully translated cue and 13 with a substitute, giving 828
language--model pairs. The substitute group has the \emph{lower} mean Prompt
Gain: $+0.8$ percentage points against $+2.8$ for the translated group, with a
median of exactly zero. Excluding the 13 raises the tier mean from $+2.1$ to
$+2.8$ points, which places Low/Zero above the Mid tier ($+2.4$) rather than
below it.

The two groups are not otherwise comparable: the substitute languages have
smaller test sets (median 50 items against 190) and lower Direct accuracy
(0.486 against 0.684, the latter well clear of the 0.5 chance level). Since a
model already at chance cannot gain from any cue, we control for Direct
accuracy. The substitute coefficient then grows from $-2.0$ to $-4.0$
percentage points ($p<0.001$, standard errors clustered by model), and the same
ordering holds separately below and above 0.6 Direct accuracy ($+2.0$ against
$+0.9$, and $+3.1$ against $+0.5$ points). It also holds in 22 of the 23
individual models. The substitute cues therefore understate the gradient
reported in \S\ref{sec:language_prompting} rather than create it.




\clearpage
\refstepcounter{section}
\label{sec:appendix_native_prompts}
\includepdf[pages=-]{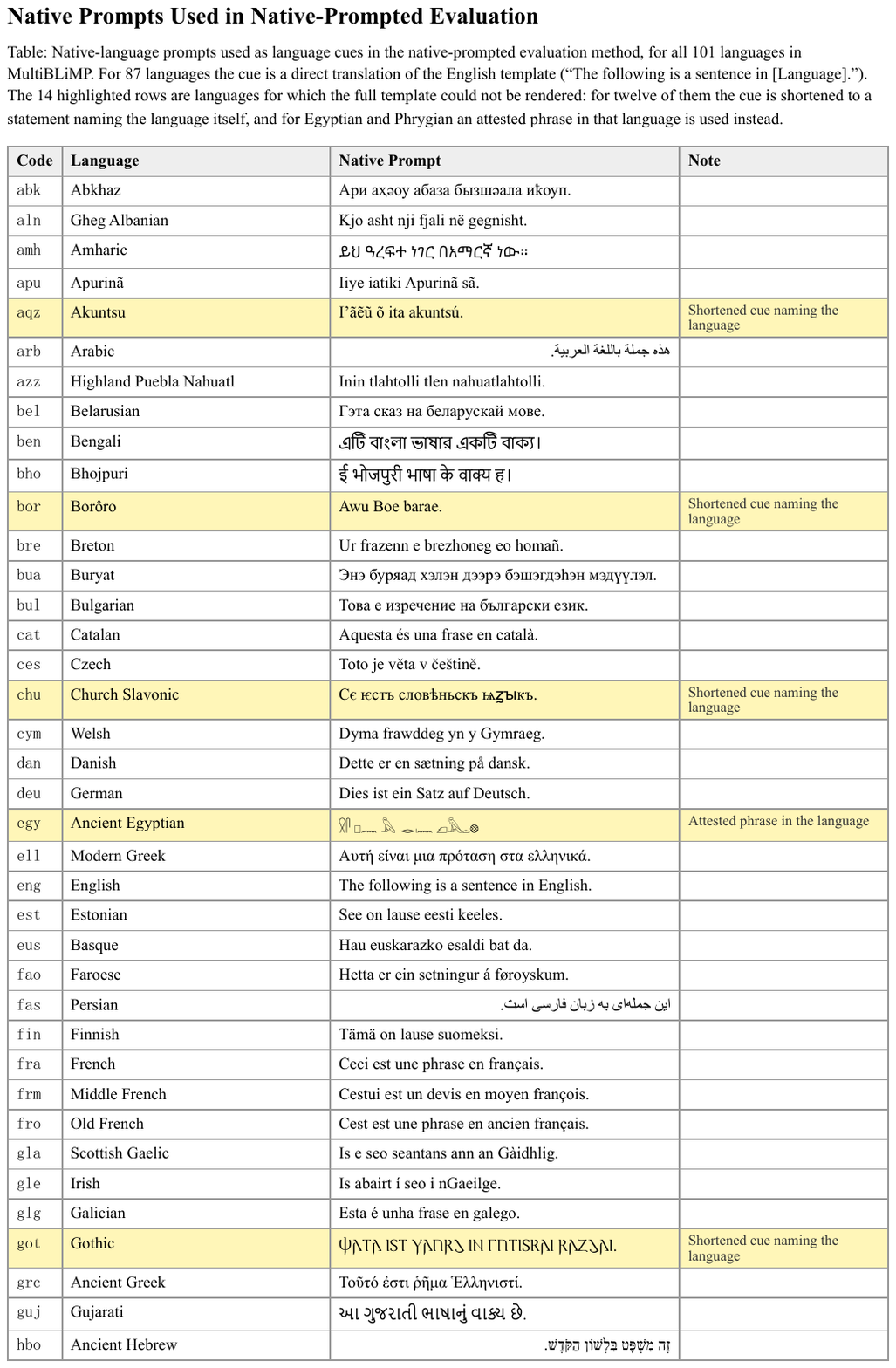}

\end{document}